%% file: main.tex
\documentclass[11pt]{article}

\usepackage[preprint]{acl}

\usepackage{times}
\usepackage{latexsym}

\usepackage[T1]{fontenc}

\usepackage[utf8]{inputenc}
\usepackage{makecell}
\usepackage{microtype}

\definecolor{commentgray}{RGB}{245,245,245}
\definecolor{commentborder}{RGB}{170,170,170}

\usepackage[most]{tcolorbox}   
\usepackage{booktabs}          
\newtcolorbox{commentbox}{
    colback=commentgray,
    colframe=commentborder,
    boxrule=0.5pt,
    arc=1pt,
    left=6pt, right=6pt, top=5pt, bottom=5pt,
    before skip=4pt, after skip=8pt
}
\usepackage{mdframed}
\newmdenv[
    topline=false,
    bottomline=false,
    rightline=false,
    linewidth=2pt,
    linecolor=gray!50,
    innerleftmargin=8pt,
    innerrightmargin=0pt,
    innertopmargin=2pt,
    innerbottommargin=2pt,
    skipabove=4pt,
    skipbelow=8pt
]{revisionbox}

\definecolor{taggray}{RGB}{125,125,125}

\usepackage{inconsolata}
\usepackage{subcaption}
\usepackage{graphicx}
\usepackage{multirow}
\usepackage[table]{xcolor}
\definecolor{ourscolor}{RGB}{220,240,205}

\usepackage{array}
\newcolumntype{C}[1]{>{\centering\arraybackslash}m{#1}}
\usepackage{booktabs}
\newcommand{\cmark}{\checkmark}
\newcommand{\xmark}{$\times$}
\newcommand{\std}[1]{\ensuremath{\mkern1mu{\scriptstyle\pm#1}}}
\usepackage{amsmath, amssymb, amsfonts}
\usepackage{booktabs}
\usepackage{natbib} 

\newif\ifbluemode
\bluemodefalse

\ifbluemode
  \definecolor{accentblue}{RGB}{0,0,255}        
  \definecolor{accentbluelight}{RGB}{220,230,248}
  \definecolor{sectionbg}{RGB}{235,240,252}
  \definecolor{reviewerbar}{RGB}{0,0,255}        
\else
  \definecolor{accentblue}{RGB}{0,0,0}
  \definecolor{accentbluelight}{RGB}{245,245,245}
  \definecolor{sectionbg}{RGB}{240,240,240}
  \definecolor{reviewerbar}{RGB}{0,0,0}
\fi
\usepackage{environ}
\newcommand{\revised}[1]{\ifbluemode\textcolor{accentblue}{#1}\else#1\fi}
\NewEnviron{revisedblock}{%
  \ifbluemode{\color{accentblue}\BODY}\else\BODY\fi}

\title{ACE: Adapter Consolidation across Experts \\  for Parameter-Efficient Fine-Tuning of MoE LLMs}

\newcommand{\system}{\textsc{ACE}}

\author{
Ahin Lee, Sehyun Yun, Joonha Park, Taesik Gong \\
Ulsan National Institute of Science and Technology (UNIST) \\
Ulsan, Republic of Korea \\
\texttt{\{ahin,nawhji,joonhapark,taesik.gong\}@unist.ac.kr}
}

\begin{document}
\maketitle




\input{camera-ready/00_abstract}
\input{camera-ready/01_intro}
\input{camera-ready/02_related_work}
\input{camera-ready/04_method}
\input{camera-ready/05_experiments}
\input{camera-ready/06_analysis}
\input{camera-ready/07_conclusion}

\input{camera-ready/08_limitation_acknowledge}


\bibliography{custom}
\appendix

\input{camera-ready/09_appendix}

\end{document}

%% file: camera-ready/00_abstract.tex
\begin{abstract}

Parameter-efficient fine-tuning (PEFT) of mixture-of-experts (MoE) models commonly attaches a separate low-rank adapter to each expert. 
This expert-wise design fragments adaptation in three ways: capacity is split across narrow low-rank updates, gradient supervision becomes sparse and imbalanced under sparse routing, and execution is decomposed into many small GEMMs. 
We find that such expert-wise separation is often unnecessary, as subsets of LoRA adapters become functionally similar during fine-tuning, revealing redundancy among expert-specific adapters.
Based on this redundancy, we propose \textbf{ACE} (Adapter Consolidation across Experts), which groups redundant experts and replaces their expert-specific adapters with group-shared higher-rank LoRA modules under the same PEFT budget.
ACE further introduces grouped adapter execution, which consolidates fragmented expert-wise adapter computations into fewer, larger group-level GEMMs.
\revised{
Across evaluations covering 12 datasets and four MoE backbones, ACE achieves the highest observed mean accuracy among the parameter-matched PEFT methods on the three backbones with complete baseline coverage, while providing $1.31\times$ to $1.48\times$ wall-clock training speedup over expert-wise LoRA without increasing peak memory. Our code is available at \url{https://github.com/UbiquitousAILab/ACE}.
}
\end{abstract}

%% file: camera-ready/01_intro.tex
\section{Introduction}

Mixture-of-Experts (MoE) models scale model capacity by activating only a small subset of experts for each input, allowing them to increase parameters without a proportional increase in per-token computation~\citep{shazeer2017outrageously, fedus2022switchtransformersscalingtrillion, du2022glam}. This efficiency has made MoE architectures increasingly popular in large language models~\citep{grattafiori2024llama, openai2024gpt4technicalreport, dai2024deepseekmoe}. However, adapting MoE models to downstream tasks remains challenging. Although sparse routing reduces per-token computation, MoE models still contain a very large number of parameters across many experts, making full fine-tuning substantially more expensive and complex than in comparable dense models~\citep{dai2024deepseekmoe, kim2021scalable, aminabadi2022deepspeed}. These challenges motivate parameter-efficient fine-tuning (PEFT) methods such as LoRA~\citep{hu2022lora}.

LoRA was originally developed for dense architectures, where it is commonly applied by attaching a separate low-rank adapter to each target module.
A direct extension to the expert-side modules of MoE models therefore attaches independent adapters to each expert~\citep{hu2022lora, meng2024pissa}. 
Although natural, this expert-wise design induces fragmentation when combined with sparse routing, along three dimensions: capacity, supervision, and execution. 
First, under a fixed trainable parameter budget, allocating separate adapters to many experts limits the rank available to each expert. 
Second, each adapter receives gradients only from tokens routed to its expert, making gradient supervision sparse and imbalanced across experts. 
Third, adapter execution is decomposed into many small 
matrix multiplication (GEMM) operations, incurring repeated launch overhead and poor hardware utilization.

\begin{figure*}[t]
    \centering
    \includegraphics[width=1\linewidth]{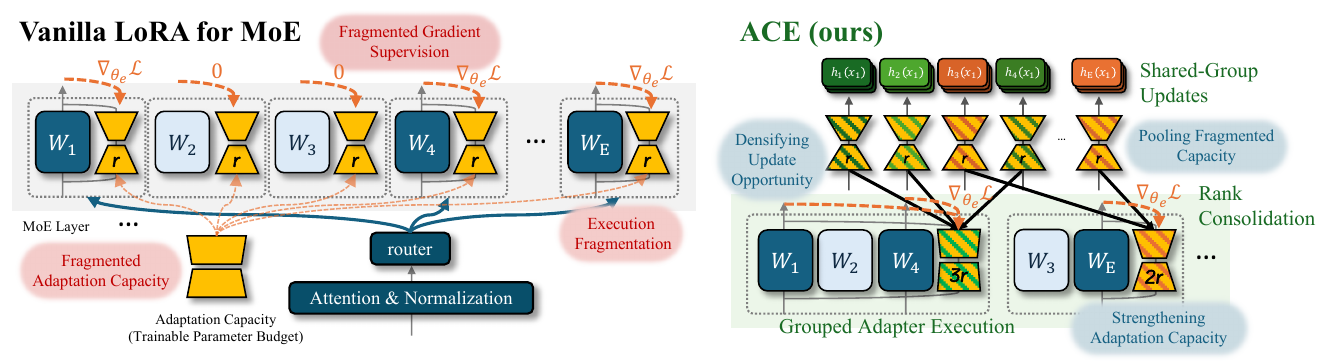}
    \caption{
    Overview of \system{}. \textbf{Left}: Standard expert-wise LoRA in MoE uses a separate rank-$r$ adapter for each expert, fragmenting adaptation capacity, yielding sparse updates, and causing many small computations. \textbf{Right}: \system{} groups experts with similar adapter-induced representations and replaces their separate adapters with group-shared higher-rank adapters under the same trainable parameter budget. Each shared adapter is updated by all experts in its group and computed once per group, improving capacity, supervision density, and execution efficiency.}
    \label{fig:ace_overview}
\end{figure*}

This fragmentation would be unavoidable if each expert required a distinct task-specific adapter. 
However, we find that expert-specific LoRA adapters are often redundant: within the same MoE layer, they form functionally similar groups during fine-tuning.
Based on this observation, we propose \textbf{\system{}} (\textbf{A}dapter \textbf{C}onsolidation across \textbf{E}xperts), a PEFT method for MoE models that consolidates functionally similar expert adapters into group-shared higher-rank LoRA modules. 
Within each MoE layer, \system{} measures adapter similarity on task-relevant inputs, groups similar experts together and replaces each group of expert-specific adapters with a shared higher-rank adapter under the same trainable parameter budget. 
This reallocation pools fragmented adaptation capacity and allows each shared adapter to receive gradients whenever any expert in its group is active. 
For execution, \system{} performs grouped adapter execution, which consolidates fragmented expert-wise adapter computations into fewer, larger group-level operations, reducing small-GEMM overhead. Figure~\ref{fig:ace_overview} provides an overview of \system{} and its key components: similarity-based grouping, rank consolidation, shared updates and grouped adapter execution.


{
\ifbluemode\color{accentblue}\fi
Across evaluations covering 12 datasets and four MoE backbones,
\system{} consistently improves adaptation quality and training
efficiency under matched parameter budgets. For example, on OLMoE,
\system{} improves the average accuracy across eight commonsense
benchmarks from 74.11\% to 75.61\% (+1.50 points) over expert-wise
LoRA, while providing 1.31$\times$--1.48$\times$ wall-clock training
speedups across the evaluated backbones without increasing peak memory.
These results indicate that similarity-guided adapter consolidation
improves both adaptation quality and execution efficiency under the
same trainable parameter budget.

}


%% file: camera-ready/02_related_work.tex
\section{Related Work}
\label{sec:related_work}

\paragraph{Parameter-Efficient Fine-Tuning.}
Parameter-efficient fine-tuning (PEFT) adapts large pretrained models by training only a small set of additional parameters while keeping the backbone frozen~\citep{houlsby2019parameter, li2021prefix, lester2021power, mahabadi2021parameter, li2024mixlora}.
Among PEFT methods, low-rank adaptation (LoRA) has become a standard approach because it parameterizes weight updates efficiently without changing the backbone architecture~\citep{hu2022lora, dettmers2023qlora}.
Subsequent work improves the use of fixed trainable parameter budgets through rank allocation or alternative low-rank parameterizations~\citep{zhang2023adalora, liu2024dora, meng2024pissa}. 
However, these methods are designed for dense architectures, where each adapted module receives gradients from all inputs. This assumption breaks down in MoE layers, where adapters are distributed across sparsely and unevenly activated experts.
Our work addresses this MoE-specific structural mismatch by studying how a fixed LoRA budget should be organized under such routing.

\paragraph{Fine-Tuning of MoE Models.}
Mixture-of-Experts (MoE) architectures scale model capacity by activating only a subset of expert networks for each input~\citep{shazeer2017outrageously, fedus2022switchtransformersscalingtrillion, du2022glam}. 
However, the large number of expert parameters increases the cost of fine-tuning for downstream adaptation.
To reduce this cost, prior work has explored selective or routed adaptation strategies for MoE models, primarily focusing on which experts or modules should be tuned. ESFT~\citep{wang2024esft} tunes the experts most relevant to a target task. PERFT~\citep{liu2025parameter} proposes PERFT-R, which applies MoE-style routing to LoRA adapters for MoE fine-tuning.
\revised{EPnG~\citep{lee2026epng} reallocates LoRA capacity across experts based on routing importance.}
In contrast, \system{} keeps the pretrained experts and router intact and addresses expert-wise LoRA fragmentation by consolidating functionally similar adapters into shared group-level modules.

%% file: camera-ready/04_method.tex
\section{Methodology}
\label{sec:method}

\subsection{Problem: Fragmentation in MoE LoRA}
\label{sec:problem}

We first identify a structural mismatch between expert-wise LoRA and sparse MoE routing. 
Consider an MoE layer with \(E\) experts, where the router activates a sparse expert subset \(\mathcal{R}(x)\) of size \(k \ll E\) for an input \(x\). 
A direct LoRA extension assigns each expert \(e\) an independent low-rank update \(\Delta W_e\), yielding an adapted expert weight \(\widetilde{W}_e = W_e + \Delta W_e\), where \(\Delta W_e = B_e A_e\). 
Let \(p_e(x)\) denote the router probability assigned to expert \(e\) for input \(x\). 
The resulting layer output is
\[
y(x)=\sum_{e\in\mathcal{R}(x)}p_e(x)(W_e x+\Delta W_e x).
\]
Although simple, this expert-wise design fragments adaptation along three coupled dimensions.

\subsubsection{Capacity Fragmentation}

Under a fixed layer-level adapter budget \(\mathcal{B}\), expert-wise LoRA divides parameters across experts. 
For expert \(e\), a rank-\(r_e\) LoRA adapter contains \((d_{\mathrm{in}}+d_{\mathrm{out}})r_e\) trainable parameters, so the layer-level budget is 
\(\mathcal{B}=\sum_{e=1}^{E}(d_{\mathrm{in}}+d_{\mathrm{out}})r_e\). 
If this budget is allocated uniformly, each expert receives
\[
r_e = \frac{\mathcal{B}}{E(d_{\mathrm{in}}+d_{\mathrm{out}})} .
\]
Thus, increasing the number of experts linearly reduces the rank available to each adapter. 
The budget is split into \(E\) narrow low-rank updates rather than forming stronger task-specific corrections.

\subsubsection{Gradient Fragmentation}

Sparse routing also fragments gradient supervision. 
For a training batch \(\mathcal{X} = \{x_n\}_{n=1}^N\), the adapter of expert \(e\) receives gradients only from tokens routed to that expert:
\[
\nabla_{\Delta W_e}\mathcal{L}
\propto
\sum_{x_n\in\mathcal{X}}
\mathbf{1}[e\in\mathcal{R}(x_n)]
p_e(x_n)\,
\nabla_{y(x_n)}\mathcal{L}\,x_n^\top .
\]
The indicator \(\mathbf{1}[e\in\mathcal{R}(x_n)]\) zeroes out contributions from all non-routed tokens.
If routing is balanced, each adapter is updated by only a \(k/E\) fraction of the batch on average. 
In practice, routing is often imbalanced~\citep{ shazeer2017outrageously,dai2024deepseekmoe}, so some adapters receive even fewer updates. 
Expert-wise LoRA therefore splits not only parameters, but also the learning signal, yielding sparse and imbalanced adapter updates.

\subsubsection{Execution Fragmentation}

Finally, expert-wise LoRA fragments execution. 
Each active adapter requires two small GEMMs,
\[
h_e=A_e x,\qquad \Delta W_e x=B_e h_e .
\]
Across a batch, these operations are invoked separately for many active experts and routed token subsets. 
This produces numerous small adapter-side GEMMs rather than fewer large matrix multiplications. 
Such small GEMMs are inefficient on GPUs because they incur repeated launch overhead and poor utilization~\citep{nvidia_cublas}.
As a result, even a small LoRA budget can introduce non-negligible runtime overhead when replicated across many experts.

These three effects arise from a common design choice in expert-wise LoRA, where each expert owns an isolated adapter while each input activates only a small subset of experts. 
This raises a central question: \emph{do expert adapters truly need to remain separate?}

\subsection{Motivation: Functional Grouping of Expert LoRA Adapters}
\label{sub:motivation}
The expert-wise separation described above is often unnecessary in practice. 
During fine-tuning, LoRA adapters within the same MoE layer frequently form functionally similar groups.
Adapters within these groups produce similar adapter-induced representations on shared task-relevant inputs. 
We quantify this behavior by computing pairwise similarity between adapter outputs on a shared probe set; details are given in Section~\ref{sub:consolidation}.

\begin{figure}[!h]
    \centering
    \includegraphics[width=1.0\linewidth]{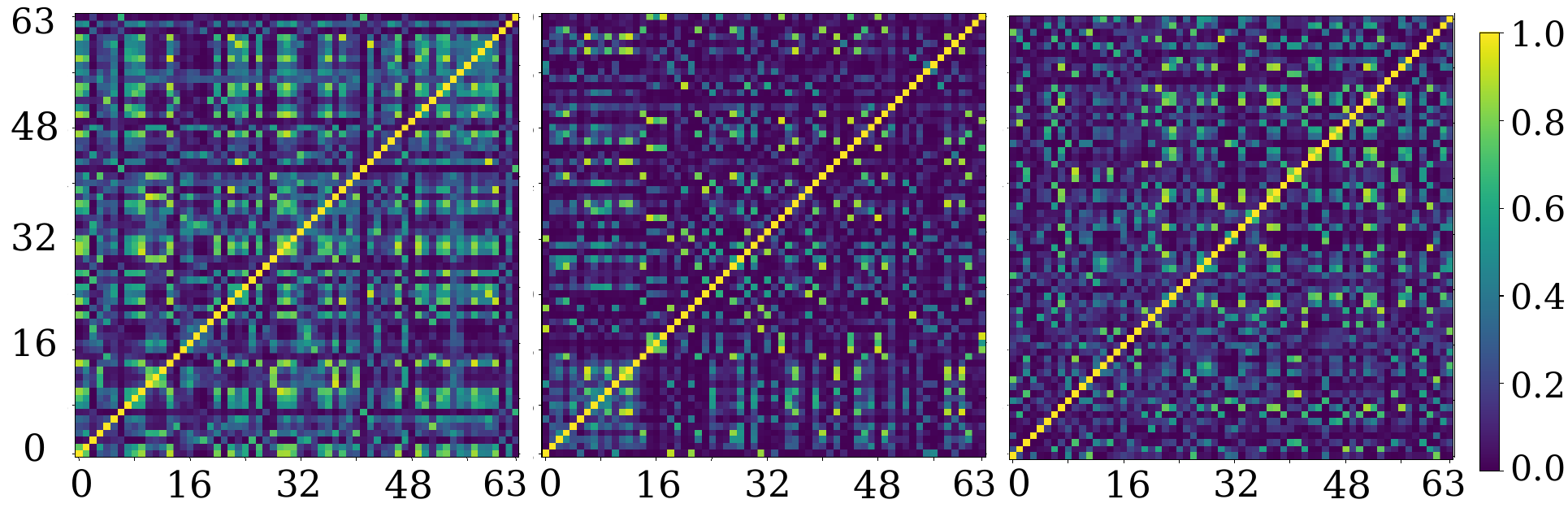}
    \caption{Functional similarity among expert LoRA adapters. 
    Each heatmap shows pairwise CKA similarity within one MoE layer, with x- and y-axes denoting expert indices. 
    Values closer to 1 indicate more similar adapter-induced representations. 
    \revised{High-similarity blocks reveal groups of experts with similar LoRA corrections. Panels correspond to layers 0, 4, and 14 of the gate projection in OLMoE on HellaSwag.}
    Full details are in Appendix~\ref{app:cka}.}
    \label{fig:similarity_matrix}
\end{figure}

Figure~\ref{fig:similarity_matrix} shows high-similarity blocks, indicating that some adapters learn similar task-specific corrections while others remain distinct. 
This motivates \system{} to consolidate functionally similar adapters into shared higher-rank modules instead of preserving many low-rank expert-specific adapters. 
\system{} uses this functional grouping structure to pool fragmented capacity via rank consolidation (Section~\ref{sub:consolidation}), densify gradient supervision via shared adapter updates (Section~\ref{sub:shared}), and reduce small-GEMM fragmentation via grouped adapter execution (Section~\ref{subsec:accel_adapter}). 

\subsection{Rank Consolidation}
\label{sub:consolidation}

\system{} first identifies functionally similar expert adapters within each MoE layer and consolidates their fragmented ranks under the same trainable parameter budget. 
After a short warm-up of \(T_{\mathrm{init}}\) steps, we collect a probe set of hidden states 
\(\mathcal{X}_{\mathrm{probe}}=\{x_n\}_{n=1}^{N}\), where \(x_n\in\mathbb{R}^{d_{\mathrm{in}}}\). 
The same probe inputs are fed to all expert adapters in the layer, regardless of their original routing decisions, so that similarity reflects adapter behavior rather than routing frequency.

For expert \(e\), we form the adapter-induced representation matrix
\[
Z_e =
\begin{bmatrix}
(\Delta W_e x_1)^\top \\
\cdots \\
(\Delta W_e x_N)^\top
\end{bmatrix}
\in \mathbb{R}^{N \times d_{\mathrm{out}}}.
\]
We mean-center each representation matrix over the probe samples as \(\bar{Z}_e = H Z_e\), where \(H = I_N - \frac{1}{N}\mathbf{1}\mathbf{1}^\top\).
For experts \(i\) and \(j\), we compute their pairwise similarity using linear CKA~\citep{kornblith2019similarity}:
\[
s_{ij}
=
\frac{\|\bar{Z}_i^\top \bar{Z}_j\|_F^2}
{\|\bar{Z}_i^\top \bar{Z}_i\|_F \, \|\bar{Z}_j^\top \bar{Z}_j\|_F}.
\]
We use CKA because \system{} groups adapters by their functional behavior rather than by raw parameter similarity. 
CKA is a standard representation-similarity measure~\citep{jiang2025tracing} and has been used in compression settings to assess representational preservation across model variants~\citep{pons2024effective}. 
Its scale invariance is useful for MoE adapters, whose output magnitudes can differ under sparse and imbalanced routing.

Using these scores, \system{} groups experts with high mutual similarity within each MoE layer.
In our experiments, we instantiate this step by constructing a weighted expert-similarity graph and applying Louvain community detection~\citep{blondel2008fast}, which avoids manually choosing a fixed similarity threshold and thus reduces sensitivity to layer-wise variation in similarity scale. 
This yields expert groups \(\{\mathcal{G}_g\}_{g=1}^{G}\).

For each group \(\mathcal{G}_g\), \system{} replaces the separate expert-specific adapters \(\{\Delta W_e\}_{e\in\mathcal{G}_g}\) with a single group-shared adapter
\[
\Delta W_g^{\mathrm{share}} = B_g A_g,
\qquad
r_g = \sum_{e\in\mathcal{G}_g} r_e ,
\]
where \(A_g\in\mathbb{R}^{r_g\times d_{\mathrm{in}}}\) and \(B_g\in\mathbb{R}^{d_{\mathrm{out}}\times r_g}\). 
Each expert \(e\in\mathcal{G}_g\) then uses
\[
\widetilde{W}_e = W_e + \Delta W_g^{\mathrm{share}}.
\]
Because experts within the same layer share input and output dimensions, this consolidation preserves the total number of trainable adapter parameters:
\[
\sum_{g=1}^{G}(d_{\mathrm{out}}+d_{\mathrm{in}})r_g
=
\sum_{e=1}^{E}(d_{\mathrm{out}}+d_{\mathrm{in}})r_e.
\]
Thus, \system{} does not increase the trainable parameter budget; it reallocates the same fragmented expert-wise ranks into fewer, higher-rank group-level adapters.

To reduce discontinuity at the consolidation step, we initialize each shared adapter from a similarity-central anchor within its group and use standard LoRA initialization for the remaining rank dimensions; details are provided in Appendix~\ref{sub:anchor}.

\subsection{Shared Adapter Updates}
\label{sub:shared}

Rank consolidation also changes how adapters receive supervision. 
After consolidation, each routed expert uses the shared adapter of its group. 
Let \(g(e)\) denote the group assignment of expert \(e\). 
The MoE output becomes
\[
y(x)
=
\sum_{e \in \mathcal{R}(x)}
p_e(x)\bigl(W_e x + \Delta W_{g(e)}^{\mathrm{share}} x\bigr).
\]

Under expert-wise LoRA, the adapter of expert \(e\) is updated only when \(e\) is routed. 
In \system{}, the shared adapter for group \(g\) is updated whenever any expert in \(\mathcal{G}_g\) is routed. 
This aggregates update opportunities at the group level, providing denser and more balanced supervision despite sparse routing.

\subsection{Grouped Adapter Execution}
\label{subsec:accel_adapter}

Rank consolidation improves adaptation capacity, but its computational benefit depends on how shared adapters are executed. 
A naïve implementation would share adapter parameters within each group while still following the expert-wise LoRA execution pattern, evaluating the same shared adapter separately for each routed expert.

\system{} avoids this by exploiting a structural property of MoE layers. 
For shared-input expert projections, such as gate and up projections, routed experts process the same input activation in parallel.
Thus, when multiple routed experts use the same group-shared adapter, the corresponding adapter output can be computed once and reused across those experts.

Formally, let \(\mathcal{R}(x)\) denote the set of experts selected by the router for input \(x\), and let
\(
\mathcal{A}(x)=\{g(e)\mid e\in\mathcal{R}(x)\}
\)
be the set of active adapter groups. 
For each active group \(g\in\mathcal{A}(x)\), \system{} evaluates
\[
c_g(x)=\Delta W_g^{\mathrm{share}}x
\]
once and reuses \(c_g(x)\) for every routed expert \(e\in\mathcal{R}(x)\) satisfying \(g(e)=g\).

Since a LoRA update consists of two GEMMs, \(h=A_gx\) and \(c_g(x)=B_gh\), this group-level evaluation replaces repeated expert-wise adapter calls with fewer, larger matrix multiplications. 
As a result, \system{} reduces small-GEMM overhead and adapter-side execution fragmentation.

%% file: camera-ready/05_experiments.tex
\section{Experiments}
\label{sec:experiments}

We evaluate \system{} on reasoning and generation tasks using MoE-based language models. 
Across all experiments, we compare methods under matched trainable parameter budgets and use the same optimization setup and evaluation protocol within each setting. 
Unless otherwise specified, models are trained with AdamW, weight decay \(0.1\).
{
\ifbluemode\color{accentblue}\fi
For \system{}, we use $T_{\mathrm{init}}=150$, a probe set of
$N=10{,}000$ tokens, and Louvain resolution $\gamma=0.8$.
}
All experiments are conducted on NVIDIA H200 GPUs.
Detailed model configurations, hyperparameters, and compute infrastructure are provided in Appendix~\ref{app:model} and Appendix~\ref{appendix:compute_infra}.

\subsection{Commonsense Reasoning}
\label{sub:commonsense}

We evaluate \system{} on eight commonsense reasoning benchmarks: HellaSwag (HS)~\citep{zellers2019hellaswag}, WinoGrande (WG)~\citep{ai2:winogrande}, ARC-Challenge (ARC-c) and ARC-Easy (ARC-e)~\citep{allenai:arc}, BoolQ~\citep{clark2019boolq}, OpenBookQA (OBQA)~\citep{OpenBookQA2018}, Physical IQA (PIQA)~\citep{Bisk2020}, and Social IQA (SIQA)~\citep{sap2019social}. 
We report accuracy for all tasks. 
We conduct experiments on three MoE-based language models: OLMoE-1B-7B~\citep{muennighoff2025olmoe}, Qwen1.5-MoE-A2.7B~\citep{qwen15moe2024}, and Moonlight~\citep{liu2025muonscalable}. 
For this setting, models are fine-tuned on Commonsense-170K and evaluated on each benchmark separately~\citep{hu2023llm}.

We compare \system{} with representative PEFT baselines, including
LoRA~\citep{hu2022lora}, AdaLoRA~\citep{zhang2023adalora},
DoRA~\citep{liu2024dora}, and PiSSA~\citep{meng2024pissa},
as well as two MoE-aware PEFT baselines, PERFT-R~\citep{liu2025parameter} and
\revised{
EPnG~\citep{lee2026epng}.}
For fair comparison, all methods adapt the same expert projections and use comparable trainable parameter budgets within each backbone.

Table~\ref{tab:main_results} reports the commonsense reasoning results. 
Across the evaluated MoE backbones, \system{} achieves the best average performance under matched trainable parameter budgets. 
The gains over dense-Transformer PEFT baselines suggest that methods designed for dense Transformer modules do not fully account for the sparse-routing structure of MoE layers, which can limit how effectively their trainable adapter parameters are utilized.
These results show that adapter consolidation provides an effective way to use limited adaptation capacity in MoE models. Results on Kimi-Linear-48B-A3B \citep{kimi2025linear}, the largest backbone evaluated, show the same trend; results are in Appendix~\ref{app:kimi_comparison}.

\begin{table*}[t]
\centering
\renewcommand{\arraystretch}{1.15}
\resizebox{\textwidth}{!}{
\begin{tabular}{l l c c c c c c c c c c}
\toprule
\textbf{Model} & \textbf{Method} & \textbf{\# Params (\%)$\downarrow$} 
& \textbf{HS}$\uparrow$ & \textbf{WG}$\uparrow$ 
& \textbf{ARC-c}$\uparrow$ & \textbf{ARC-e}$\uparrow$ 
& \textbf{BoolQ}$\uparrow$ & \textbf{OBQA}$\uparrow$ & \textbf{PIQA}$\uparrow$
& \textbf{SIQA}$\uparrow$ & \textbf{Avg.}$\uparrow$ \\
\midrule

\multirow{7}{*}{\shortstack{\textbf{OLMoE}\\{ (1B/7B)}}}
& LoRA        & 0.27 & 90.82\std{0.23} & 70.08\std{0.34} & 69.42\std{1.00} & 85.61\std{0.09} & 66.71\std{1.12} & 74.60\std{0.92} & 83.10\std{0.22} & 52.51\std{0.90} & 74.11\std{0.39} \\
& DoRA        & 0.33 & 90.98\std{0.27} & 70.25\std{0.34} & 69.46\std{0.31} & 85.98\std{0.14} & 66.96\std{0.71} & 74.93\std{1.21} & 83.31\std{0.30} & 52.35\std{0.89} & 74.28\std{0.23} \\
& AdaLoRA     & 0.54 $\rightarrow$ 0.27 & 90.68\std{0.14} & 68.14\std{0.25} & 68.51\std{0.23} & 84.69\std{0.84} & 65.52\std{0.39} & 73.67\std{0.12} & 83.08\std{0.47} & 51.31\std{0.61} & 73.20\std{0.11} \\
& PiSSA       & 0.27 & 90.98\std{0.19} & 70.46\std{0.25} & 69.96\std{0.44} & \textbf{86.38}\std{0.27} & \textbf{68.71}\std{0.34} & 75.27\std{0.50} & 83.44\std{0.55} & 52.83\std{0.53} & 74.75\std{0.16} \\
& {PERFT-R}     & {0.27} & {87.36\std{0.89}} & {71.46\std{2.43}} & {66.38\std{1.13}} & {82.76\std{0.28}} & {64.04\std{4.73}} & {73.87\std{0.23}} & {80.05\std{0.85}} & {51.62\std{0.10}} & {72.19\std{0.49}} \\
& {EPnG}
& {0.27}
& {90.82\std{0.34}}
& {68.95\std{0.55}}
& {68.83\std{0.60}}
& {85.75\std{0.60}}
& {66.96\std{0.52}}
& {74.47\std{1.40}}
& {83.15\std{0.39}}
& {52.27\std{0.21}}
& {73.90\std{0.33}} \\
& \cellcolor{ourscolor}\textbf{\system{} (ours)}
& \cellcolor{ourscolor}0.27
& \cellcolor{ourscolor}\textbf{92.24}\std{0.03}
& \cellcolor{ourscolor}\textbf{72.85}\std{0.51}
& \cellcolor{ourscolor}\textbf{70.88}\std{0.18}
& \cellcolor{ourscolor}86.21\std{0.34}
& \cellcolor{ourscolor}68.04\std{1.99}
& \cellcolor{ourscolor}\textbf{77.33}\std{1.10}
& \cellcolor{ourscolor}\textbf{83.93}\std{0.14}
& \cellcolor{ourscolor}\textbf{53.41}\std{0.44}
& \cellcolor{ourscolor}\textbf{75.61}\std{0.28} \\

\midrule

\multirow{7}{*}{\shortstack{\textbf{Qwen}\\{ (3B/14B)}}}
& LoRA        & 0.21 & 93.23\std{0.24} & 77.66\std{0.21} & 79.44\std{0.30} & 90.63\std{0.07} & 70.84\std{0.39} & 82.67\std{0.58} & \textbf{86.80}\std{0.35} & 56.72\std{0.30} & 79.75\std{0.03} \\
& DoRA        & 0.26 & 93.25\std{0.25} & 78.08\std{0.25} & \textbf{79.64}\std{0.47} & 90.71\std{0.12} & 70.58\std{0.51} & 82.80\std{1.11} & 86.73\std{0.20} & 56.64\std{0.30} & 79.80\std{0.19} \\
& AdaLoRA     & 0.42 $\rightarrow$ 0.21 & 93.21\std{0.07} & 74.45\std{0.30} & 78.89\std{0.13} & 90.42\std{0.11} & 69.49\std{1.10} & 82.07\std{0.50} & 86.29\std{0.30} & 55.63\std{0.67} & 78.81\std{0.22} \\
& PiSSA       & 0.21 & 93.28\std{0.18} & \textbf{79.19}\std{0.67} & 78.79\std{0.54} & 90.69\std{0.44} & \textbf{71.70}\std{0.11} & 83.60\std{0.53} & 86.58\std{0.22} & 56.05\std{0.26} & 79.98\std{0.11} \\
& {PERFT-R} & {0.21} & {92.02\std{0.32}} & {73.35\std{0.73}} & {78.38\std{0.3}} & {\textbf{90.95\std{0.19}}} & {69.45\std{0.88}} & {80.93\std{0.42}} & {85.95\std{0.36}} & {53.91\std{0.77}} & {78.12\std{0.20}} \\
& {EPnG}
& {0.21}
& {93.50\std{0.08}}
& {76.53\std{0.48}}
& {79.78\std{0.39}}
& {90.67\std{0.16}}
& {69.46\std{1.23}}
& {83.40\std{0.20}}
& {86.38\std{0.08}}
& {56.64\std{0.26}}
& {79.54\std{0.09}} \\
& \cellcolor{ourscolor}\textbf{\system{} (ours)}
& \cellcolor{ourscolor}0.21
& \cellcolor{ourscolor}\textbf{93.81}\std{0.17}
& \cellcolor{ourscolor}\textbf{79.19}\std{0.56}
& \cellcolor{ourscolor}79.50\std{0.35}
& \cellcolor{ourscolor}90.66\std{0.22}
& \cellcolor{ourscolor}71.49\std{0.47}
& \cellcolor{ourscolor}\textbf{83.87}\std{1.15}
& \cellcolor{ourscolor}86.67\std{0.62}
& \cellcolor{ourscolor}\textbf{56.99}\std{0.55}
& \cellcolor{ourscolor}\textbf{80.27}\std{0.18} \\

\midrule

\multirow{7}{*}{\shortstack{\textbf{Moonlight}\\{ (3B/16B)}}}
& LoRA        & 0.22 & 92.23\std{0.07} & 74.59\std{0.72} & 84.67\std{0.27} & 94.28\std{0.11} & 67.17\std{0.94} & 82.00\std{0.72} & 86.92\std{0.36} & 52.47\std{0.92} & 79.29\std{0.32} \\
& DoRA        & 0.27 & 92.15\std{0.14} & 74.67\std{0.36} & 84.93\std{0.47} & 94.35\std{0.18} & 67.59\std{0.26} & 82.07\std{0.83} & 86.98\std{0.55} & 52.29\std{0.50} & 79.38\std{0.30} \\
& AdaLoRA     & 0.43 $\rightarrow$ 0.22 & 91.63\std{0.15} & 72.64\std{0.45} & \textbf{84.96}\std{0.10} & 94.08\std{0.32} & 67.52\std{0.42} & 81.16\std{0.32} & 86.74\std{0.08} & 51.93\std{0.55} & 78.83\std{0.18} \\
& PiSSA       & 0.22 & 91.38\std{0.12} & 72.45\std{0.16} & 84.67\std{0.18} & 93.93\std{0.15} & 62.53\std{2.38} & 80.80\std{0.53} & 86.83\std{0.17} & 52.25\std{0.71} & 78.11\std{0.23} \\
& {PERFT-R}  & {0.22} & {90.9\std{0.13}} & {72.53\std{0.75}} & {84.13\std{0.39}} & {94.05\std{ 0.13}} & {65.92\std{ 1.4}} & {81.27\std{ 0.9}} & {86.71\std{ 0.39}} & {51.40\std{ 0.47}} & {78.36\std{0.51}} \\
& {EPnG}
& {0.22}
& {92.95\std{0.05}}
& {77.03\std{0.57}}
& {84.78\std{0.44}}
& {94.50\std{0.38}}
& {67.56\std{0.35}}
& {83.53\std{0.64}}
& {87.69\std{0.59}}
& {52.97\std{0.75}}
& {80.13\std{0.24}} \\
& \cellcolor{ourscolor}\textbf{\system{} (ours)}
& \cellcolor{ourscolor}0.22
& \cellcolor{ourscolor}\textbf{93.49}\std{0.30}
& \cellcolor{ourscolor}\textbf{79.95}\std{0.34}
& \cellcolor{ourscolor}84.30\std{0.09}
& \cellcolor{ourscolor}\textbf{94.75}\std{0.21}
& \cellcolor{ourscolor}\textbf{69.31}\std{0.82}
& \cellcolor{ourscolor}\textbf{84.93}\std{1.01}
& \cellcolor{ourscolor}\textbf{87.71}\std{0.38}
& \cellcolor{ourscolor}\textbf{54.93}\std{0.26}
& \cellcolor{ourscolor}\textbf{81.17}\std{0.19} \\

\bottomrule
\end{tabular}
}
\caption{Commonsense reasoning results across MoE backbones. 
Model entries show active/total backbone parameters. 
The \# Params column denotes the percentage of trainable adapter parameters. 
We report accuracy for each benchmark; the Avg. column reports the mean benchmark accuracy averaged {over three runs}, with standard deviation across runs. 
The best result for each model and metric is shown in bold.}
\label{tab:main_results}
\end{table*}

\begin{table}[t]
\centering
\small
\setlength{\tabcolsep}{4pt}
\renewcommand{\arraystretch}{1.15}
\begin{tabular}{lccccc}
\toprule
\multirow{2}{*}{\shortstack{\\\\\textbf{Model}}} 
& \multicolumn{2}{c}{\textbf{Peak Mem. (GB)}} 
& \multicolumn{2}{c}{\textbf{Train Time (hr)}} 
& \multirow{2}{*}{\shortstack{\\\\\textbf{Speed}\\\textbf{up}}} \\
\cmidrule(lr){2-3} \cmidrule(lr){4-5}
& \textbf{LoRA} & \textbf{\system{}} 
& \textbf{LoRA} & \textbf{\system{}} 
& \\
\midrule
OLMoE       & 98.21  & 98.18  & 2.47  & \textbf{1.89}  & 1.31$\times$ \\
Qwen        & 89.90  & 89.84  & 6.18  & \textbf{4.19}  & 1.48$\times$ \\
Moonlight   & 48.80  & 48.80  & 23.75 & \textbf{16.77} & 1.42$\times$ \\
Kimi & 118.22 & 118.22 & 87.39 & \textbf{60.41} & 1.45$\times$ \\
\bottomrule
\end{tabular}
\caption{
Training efficiency over three epochs. 
We compare \system{} with expert-wise LoRA under the same batch size and training setup.
Speedup is computed as LoRA time divided by \system{} time.}
\label{tab:training_efficiency}
\end{table}

\subsection{Mathematical Reasoning, Code Generation, and Legal Understanding}
\label{sub:math_code_legal}
To test whether the gains of \system{} extend beyond commonsense reasoning, we evaluate four domain-specific benchmarks on OLMoE-1B-7B and Qwen1.5-MoE-A2.7B.
We fine-tune on MetaMathQA~\citep{yu2024metamath} for GSM8K~\citep{cobbe2021gsm8k} and MATH~\citep{hendrycksmath2021}, on CodeFeedback-Filtered-Instruction~\citep{zheng2024opencodeinterpreter} for HumanEval~\citep{chen2021evaluating}, and use CaseHOLD~\citep{zheng2021casehold} for legal understanding.
Training and evaluation details are in Appendix~\ref{app:math_code_details}.

Table~\ref{tab:math_code_legal_results} shows that \system{} achieves the best average on both backbones and leads on most individual benchmarks, indicating that adapter consolidation generalizes across specialized domains.



\subsection{Computational and Memory Efficiency}
\label{sub:efficiency}
We evaluate computational efficiency against expert-wise LoRA, the execution pattern directly targeted by \system{}. 
All methods are compared under the same effective batch size and training setup within each backbone. 
As shown in Table~\ref{tab:training_efficiency}, \system{} reduces wall-clock training time across all evaluated MoE backbones, with speedups ranging from 1.31$\times$ on OLMoE to 1.48$\times$ on Qwen, 1.42$\times$ on Moonlight, and 1.45$\times$ on Kimi. 
The improvement comes without additional memory cost: peak memory remains unchanged compared with LoRA.

These results indicate that grouped adapter execution effectively reduces the small-GEMM overhead introduced by expert-wise LoRA. 
Instead of repeatedly evaluating separate adapter computations for each routed expert, \system{} evaluates shared adapters at the group level, yielding practical training-time gains while preserving the same trainable parameter budget. 
Additional runtime and memory results for other PEFT baselines are provided in Appendix~\ref{app:efficiency}, and a per-stage breakdown of the consolidation pipeline in Appendix~\ref{app:overhead_breakdown}.

\setlength{\tabcolsep}{4.2pt}
\renewcommand{\arraystretch}{1.05}
\begin{table*}[t]
\centering
\resizebox{\textwidth}{!}{
\begin{tabular}{l c c c c c c c c c c c c}
\toprule
& \multicolumn{6}{c}{\textbf{OLMoE}} 
& \multicolumn{6}{c}{\textbf{Qwen}} \\
\cmidrule(lr){2-7} \cmidrule(lr){8-13}
\textbf{Method}
& \textbf{\# Params (\%)$\downarrow$}
& \textbf{GSM8K}$\uparrow$
& \textbf{MATH}$\uparrow$
& \textbf{HEval}$\uparrow$
& \textbf{{CHold}}$\uparrow$
& \textbf{Avg.}$\uparrow$
& \textbf{\# Params (\%)$\downarrow$}
& \textbf{GSM8K}$\uparrow$
& \textbf{MATH}$\uparrow$
& \textbf{HEval}$\uparrow$
& \textbf{{CHold}}$\uparrow$
& \textbf{Avg.}$\uparrow$ \\
\midrule
LoRA
& 1.08 & 65.73 & 25.95 & 30.49 & {78.53} & 50.18
& 0.86 & 66.34 & 36.31 & 77.44 & {72.11} & 63.05 \\
AdaLoRA
& 2.16$\rightarrow$1.08 & 65.20 & 24.68 & 26.22 & {75.65} & 47.94
& 1.72$\rightarrow$0.86 & 66.64 & 38.33 & 79.27 & {69.06} & 63.33 \\
DoRA
& 1.14 & 66.11 & 25.78 & 28.05 & {78.64} & 49.65
& 0.91 & 67.10 & 37.41 & 77.44 & {72.26} & 63.55 \\
PiSSA
& 1.08 & 65.88 & 24.68 & \textbf{32.32} & {79.83} & 50.68
& 0.86 & 66.19 & \textbf{39.34} & 75.61 & {74.93} & 64.02 \\
\rowcolor{ourscolor}
\textbf{\system{} (ours)}
& 1.08 & \textbf{66.49} & \textbf{26.96} & 31.10 & \textbf{{82.57}} & \textbf{51.78}
& 0.86 & \textbf{67.48} & 37.49 & \textbf{79.27} & \textbf{{75.24}} & \textbf{64.87} \\
\bottomrule
\end{tabular}
    }
\caption{Results on mathematical reasoning, code generation, and legal understanding across MoE backbones.
We report accuracy for GSM8K, MATH, and {CaseHOLD (CHold)}, and pass@10 for HumanEval (HEval).
The Avg.\ reports the mean across the four benchmarks.
The best result for each model and metric is shown in bold.}
\label{tab:math_code_legal_results}
\end{table*}

%% file: camera-ready/06_analysis.tex
\section{Further Analysis}  
\label{sec:analysis}
{

\subsection{Does \system{} Mitigate MoE LoRA Fragmentations?}
\label{sec:fragmentation_mitigation}

We evaluate whether \system{} mitigates the three fragmentations from Section~\ref{sec:problem} using effective rank (capacity), gradient statistics (supervision), and CUDA profiling (execution).

}

\paragraph{Capacity fragmentation.}
We measure the \emph{effective rank} of each learned adapter update 
\(\Delta W = BA\), computed as the exponential of the entropy of its 
normalized singular values~\citep{roy2007effective}. 
This metric captures how much of the nominal low-rank capacity is effectively used by the learned update.

\begin{figure}[!h]
    \centering
    \includegraphics[width=1\linewidth]{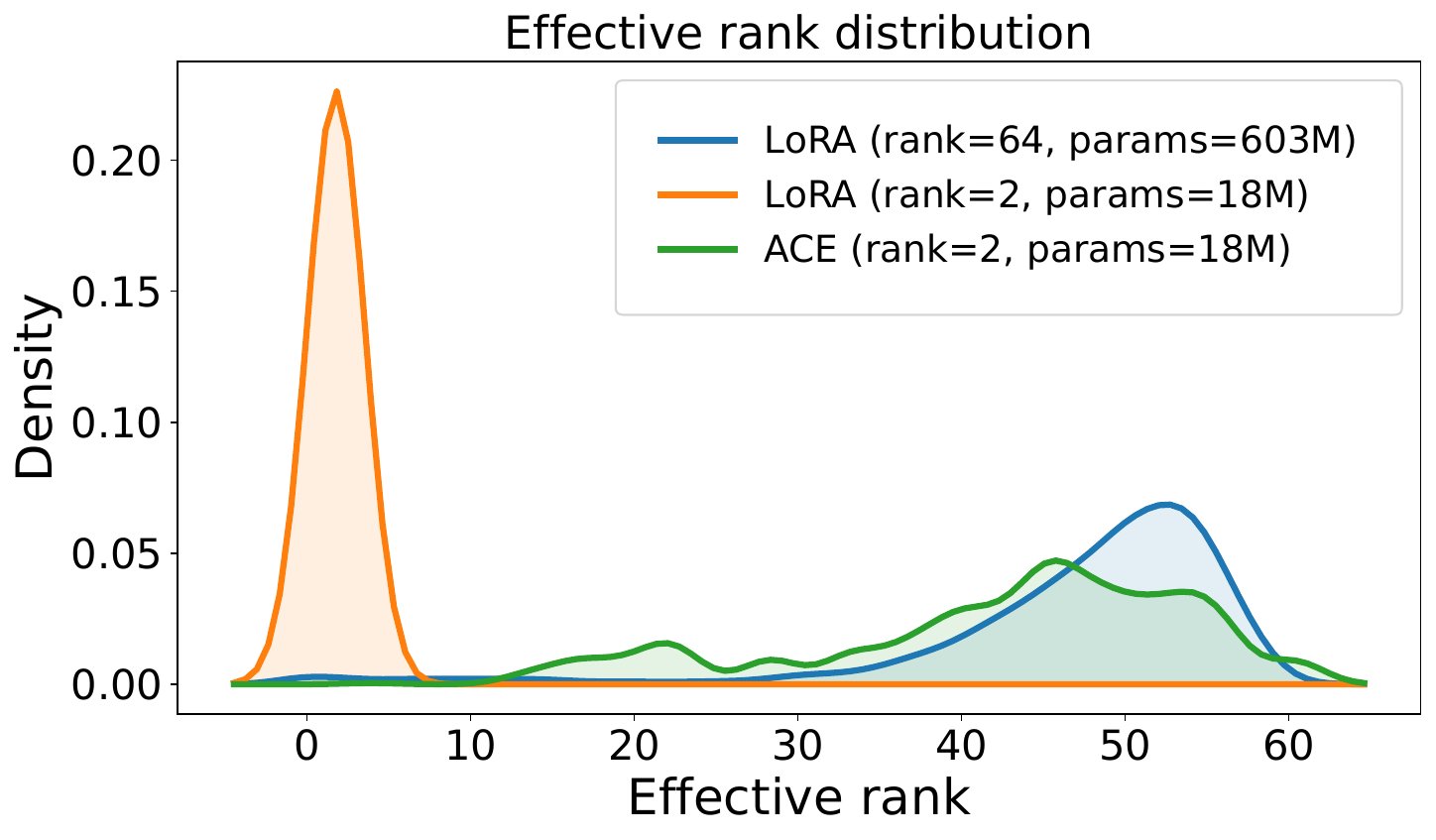}
    \caption{Distribution of effective ranks for LoRA-\(r{=}2\), LoRA-\(r{=}64\), and \system{}. 
    The parameter counts denote the total number of trainable parameters for each method.}
    \label{fig:effective_rank_distribution}
\end{figure}

Figure~\ref{fig:effective_rank_distribution} compares the effective-rank distributions of LoRA-\(r{=}2\), LoRA-\(r{=}64\), and \system{} with base rank \(r{=}2\). 
LoRA-\(r{=}2\) is strongly bottlenecked
(mean effective rank 1.77),
whereas LoRA-\(r{=}64\) shifts the distribution to a mean of 46.8 at the cost of roughly \(32\times\) more trainable parameters. 
Under the same budget as LoRA-\(r{=}2\), \system{} shifts the distribution close to LoRA-\(r{=}64\), indicating that pooling expert-wise low-rank budgets recovers effective adaptation capacity without increasing trainable parameters.

{
\paragraph{Gradient-supervision fragmentation.}
\begin{figure}[t]
\centering
\includegraphics[width=0.49\columnwidth]{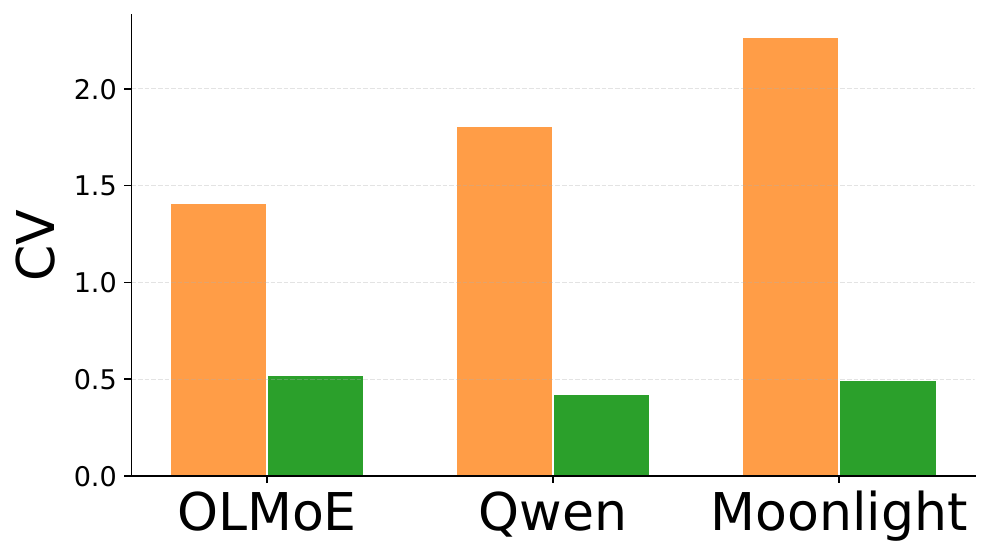}
\includegraphics[width=0.49\columnwidth]{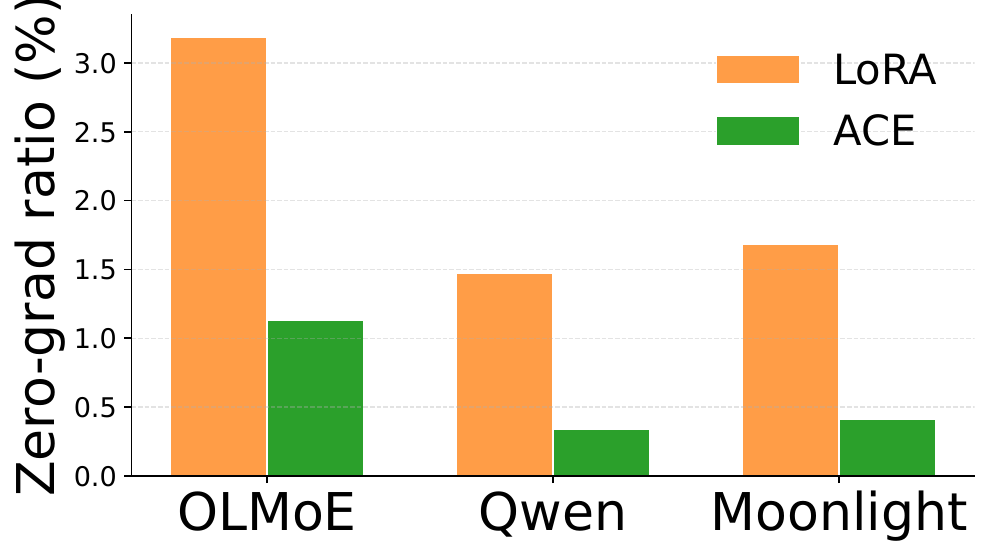}\hfill
\caption{
Gradient-supervision fragmentation for up-projection LoRA adapters across three MoE backbones. 
\system{} reduces both gradient-norm imbalance and the zero-gradient ratio relative to expert-wise LoRA. 
}
\label{fig:grad_frag}
\end{figure}
Under sparse routing, each expert-wise LoRA receives gradients only when its expert is selected, producing uneven and often near-zero updates. 
To quantify this, we measure two statistics over per-expert gradient norms within each MoE layer: 
(i) the coefficient of variation (CV), capturing imbalance across experts, and 
(ii) the fraction of adapter rows with entirely zero gradients, capturing the prevalence of unsupervised parameters. 
Both are averaged over 50 training steps. Additional measurement details and gate-projection results are provided in Appendix~\ref{app:grad_gate}.
Figure~\ref{fig:grad_frag} shows that \system{} consistently reduces both metrics across Qwen, OLMoE, and Moonlight backbones. 
For instance, on OLMoE \texttt{up\_proj}, \system{} reduces the CV from 1.449 to 0.531 and the zero-gradient ratio from 3.18\% to 1.17\%. 
These results indicate that shared-group updates densify supervision by updating each consolidated adapter whenever any expert in its group is routed.

\paragraph{Execution fragmentation.}

\begin{table}[t]
\centering
\small
\setlength{\tabcolsep}{4pt}
\renewcommand{\arraystretch}{1.12}
\resizebox{\columnwidth}{!}{%
\begin{tabular}{lcc}
\toprule
\textbf{Metric} & \textbf{LoRA} & \textbf{\system{}} \\
\midrule
Matrix multiplication calls    & 106.0K & 62.2K ($-41.3\%$) \\
Matrix multiplication CUDA time & 838ms  & 558ms ($-33.4\%$) \\
CUDA kernel events             & 482.9K & 389.0K ($-19.4\%$) \\
CUDA kernel time               & 10.36s & 8.14s ($-21.4\%$) \\
Wall-clock time                & 20.24s & 17.68s ($-12.6\%$) \\
\bottomrule
\end{tabular}
}
\caption{
CUDA profiling prefill using \texttt{torch.profiler}. 
Matrix multiplication rows correspond to PyTorch \texttt{aten::mm} operations in the profiled region. 
Kernel-level rows include all CUDA operations in the same region.
Values in parentheses indicate the relative reduction compared with LoRA.}
\label{tab:cuda_profile}
\end{table}

For execution fragmentation, we examine whether grouping adapters reduces the many small matrix multiplications induced by expert-wise LoRA. 
We profile prefill on Qwen1.5-MoE-A2.7B, measure matrix multiplication calls, their CUDA execution time, and overall CUDA kernel activity. 
Table~\ref{tab:cuda_profile} shows that \system{} consistently reduces these profiler-level indicators, including matrix multiplication calls, CUDA kernel events, and CUDA execution time. 
The reduction also translates into lower wall-clock prefill time, although the wall-clock gain is smaller because the profiled region includes non-adapter computation, CPU-side overhead, and framework overhead. 
These results indicate that \system{} reduces execution fragmentation in practice.
}

\subsection{CKA Similarity Reflects Functional Redundancy}
\label{sec:expert_redundancy}

\begin{figure}[ht]
    \centering
    \includegraphics[width=\linewidth]{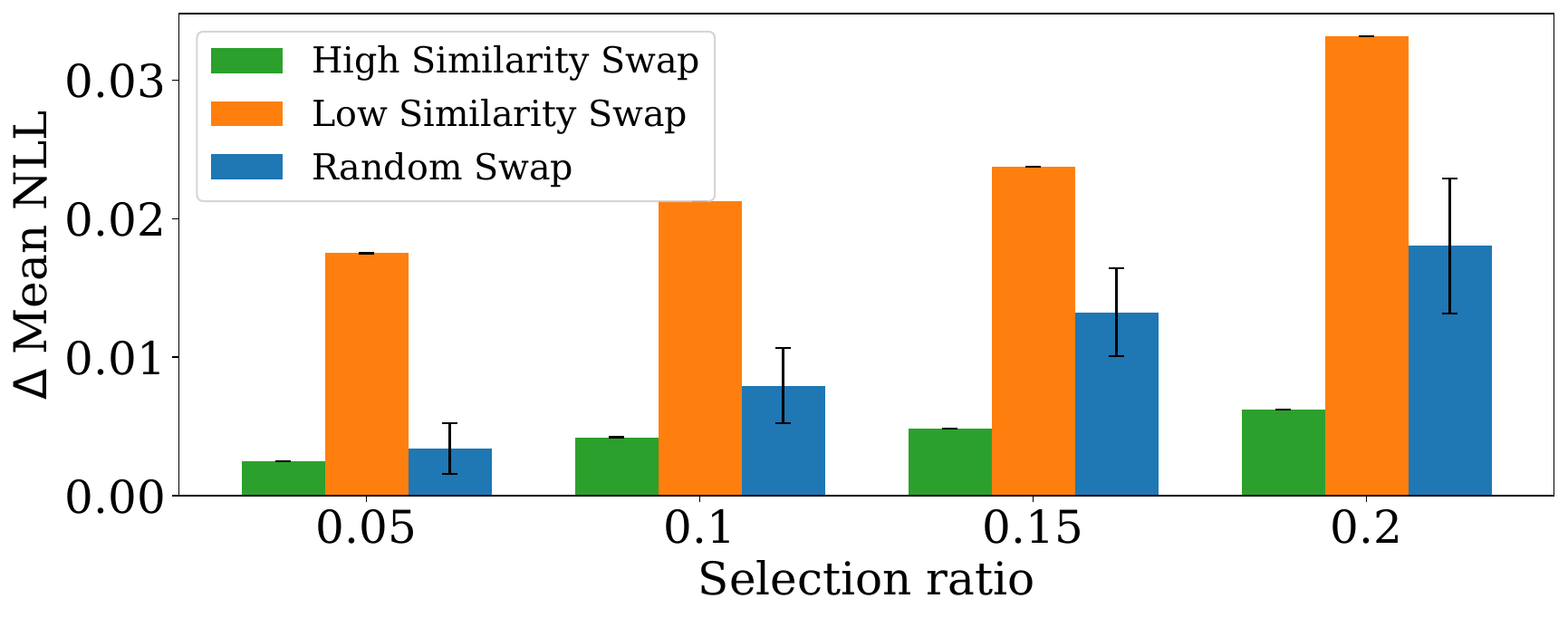}
    \caption{Adapter swapping analysis. 
    Swapping high-CKA adapters causes much smaller NLL degradation than swapping low-CKA or random adapters, indicating that CKA similarity reflects functional interchangeability.}
    \label{fig:expert_redundancy}
\end{figure}

\system{} groups experts whose adapter representations are similar under CKA. 
To justify this criterion, we test whether high-CKA similarity implies that two adapters make similar task-specific corrections. 
We use adapter swapping as a direct test: if two adapters are functionally redundant, replacing one with the other without retraining should cause only a small change in model behavior.

For each MoE layer, we choose an anchor gate-projection adapter, rank the remaining adapters by CKA similarity, and replace either the top-\(\rho\), bottom-\(\rho\), or random-\(\rho\) fraction with the anchor, where \(\rho \in \{0.05,0.10,0.15,0.20\}\). 
We then measure the change in negative log-likelihood, \(\Delta\mathrm{NLL}\), without retraining.

Figure~\ref{fig:expert_redundancy} shows that high-similarity swaps cause substantially smaller NLL degradation than low-similarity or random swaps across selection ratios. 
This indicates that CKA similarity captures functional interchangeability, meaning that CKA-based grouping can consolidate adapters while minimizing behavioral disruption.
Up-projection results are provided in Appendix~\ref{app:swapping_up_proj}.

We further observe positive CKA gaps across all four evaluated backbones
on commonsense reasoning (0.298--0.438), as well as across mathematical
reasoning, code generation, and legal understanding on OLMoE and Qwen.
This indicates that the groupable functional structure exploited by
\system{} generalizes across both model architectures and task domains.
Details are provided in Appendix~\ref{app:cka_cross_task}.

\begin{table}[t]
\centering
\small
\setlength{\tabcolsep}{4pt}
\renewcommand{\arraystretch}{1.12}
\resizebox{\columnwidth}{!}{%
\begin{tabular}{lccc ccc}
\toprule
\textbf{Method}
& \shortstack{\textbf{Sim.}\\\textbf{Group}}
& \shortstack{\textbf{Rank}\\\textbf{Cons.}}
& \shortstack{\textbf{Grp.}\\\textbf{Exec.}}
& \textbf{Avg.}$\uparrow$
& \textbf{Time}$\downarrow$
& \textbf{Mem.}$\downarrow$ \\
\midrule
LoRA
& \xmark & \xmark & \xmark
& 74.11
& 2.47
& 98.21 \\
\midrule
\multicolumn{7}{l}{\textit{w/o Sim. Group}} \\
\quad Random
& \xmark & \cmark & \cmark
& 74.72 {\scriptsize $(-0.89)$}
& 1.89
& 98.18 \\
\quad Full Merge
& \xmark & \cmark & \cmark
& 73.57 {\scriptsize $(-2.04)$}
& 2.03 {\scriptsize $(+0.14)$}
& 97.61 \\
w/o Rank Cons.
& \cmark & \xmark & \cmark
& 72.63 {\scriptsize $(-2.98)$}
& 1.84
& 96.54 \\
w/o Grp. Exec.
& \cmark & \cmark & \xmark
& 75.61
& 2.83 {\scriptsize $(+0.94)$}
& 98.18 \\
\midrule
\textbf{\system{}}
& \cmark & \cmark & \cmark
& {75.61}
& {1.89}
& {98.18} \\
\bottomrule
\end{tabular}%
}
\caption{
Component ablation of \system{}. 
\emph{Sim.\ Group} denotes CKA-based functional grouping; 
\emph{Rank cons.} pools per-expert ranks into a higher-rank group adapter; 
\emph{Grp.\ exec.} evaluates each shared adapter once per active group.
Under \emph{w/o Sim. Group}, \emph{Random} groups experts randomly, 
and \emph{Full Merge} consolidates all experts into one group.
Time in hours, memory in GB.}
\label{tab:ace_component_ablationx}
\end{table}

{
\subsection{Ablation Study}
\label{sub:ablation_study}
Table~\ref{tab:ace_component_ablationx} ablates the main components of \system{} on OLMoE. 
\textit{Full Merge} performs worse than LoRA, showing that excessive consolidation can remove useful expert specialization, while Random Group removes the CKA-based alignment used by \system{}. Among the variants that preserve the trainable-parameter budget, \textit{Full Merge} has the lowest peak memory (97.61 GB) since grouped execution caches one activation per group, confirming that consolidation reduces activation memory alongside compute.
\textit{No Rank Cons.} isolates the effect of sharing without pooling expert-wise ranks. 
\textit{No Grp. Exec.} keeps the same consolidated adapters as \system{} but disables grouped execution; accuracy is unchanged, while runtime increases from 1.89 to 2.83 hours. 

These results show that the components play distinct roles: similarity grouping selects which adapters to consolidate, rank consolidation increases effective capacity, and grouped execution provides the runtime gain.

\subsection{Expert Adaptation Is a Primary Target in MoE PEFT}
\label{sec:expert_importance}

\begin{table}[t]
\centering
\footnotesize
\setlength{\tabcolsep}{3pt}
\renewcommand{\arraystretch}{1.15}

\resizebox{\columnwidth}{!}{%
\begin{tabular}{lcccc}
\toprule
& \multicolumn{2}{c}{\textbf{OLMoE}}
& \multicolumn{2}{c}{\textbf{Qwen}} \\
\cmidrule(lr){2-3}\cmidrule(lr){4-5}

\textbf{Adaptation Scope}
& \textbf{\# Params} & \textbf{Avg.} $\uparrow$
& \textbf{\# Params} & \textbf{Avg.} $\uparrow$ \\
\midrule

Attention only (Q, K, V)
& 0.36 & $71.46 \std{0.20}$
& 0.26 & $79.72 \std{0.20}$ \\
\midrule

MoE only (LoRA)
& 0.27 & $74.11 \std{0.39}$
& 0.21 & $79.75 \std{0.03}$ \\

MoE only (\system{})
& 0.27 & $\mathbf{75.61} \std{0.28}$
& 0.21 & $\mathbf{80.27} \std{0.18}$ \\
\midrule

Attn + MoE (LoRA)
& 0.63 & $75.12 \std{0.38}$
& 0.47 & $80.46 \std{0.13}$ \\

Attn + MoE (\system{})
& 0.63 & $\mathbf{75.99} \std{0.16}$
& 0.47 & $\mathbf{80.59} \std{0.12}$ \\
\bottomrule
\end{tabular}%
}

\caption{
Effect of adaptation scope on OLMoE and Qwen,
averaged over eight commonsense reasoning benchmarks and three runs.
The \# Params columns report the percentage of trainable parameters.}
\label{tab:adaptation_scope}
\end{table}

Finally, we examine whether expert-side adaptation is necessary, or whether one can avoid expert-wise fragmentation by adapting only attention modules. 
To this end, Table~\ref{tab:adaptation_scope} compares attention-only, MoE-only, and combined attention-plus-MoE adaptation on OLMoE and Qwen.

Across both backbones, MoE-side adaptation provides a stronger
performance--parameter trade-off than attention-only adaptation.
In particular, MoE-only \system{} outperforms attention-only adaptation
while using fewer trainable parameters on both OLMoE and Qwen.
This trend is consistent with recent findings that MLP-side adaptation
can be more effective than attention-side adaptation in Transformer
fine-tuning~\citep{hayou2026plop, ward2025rank}.

When attention and MoE adapters are trained together, \system{} again improves over the LoRA counterpart. 
This suggests that adapter consolidation is complementary to attention-side adaptation rather than merely replacing it. 
Overall, these results support our focus on expert-side LoRA fragmentation as a central issue in MoE PEFT.
}

%% file: camera-ready/07_conclusion.tex
\section{Conclusion}
\label{sec:conclusion}

We introduced \system{}, a PEFT method that addresses the three fragmentations of expert-wise LoRA in MoE models: capacity, gradient supervision, and execution. By consolidating functionally aligned expert adapters into group-shared higher-rank modules with grouped execution, \system{} achieves a stronger quality-efficiency trade-off than parameter-matched PEFT baselines across multiple MoE backbones and benchmarks. 
More broadly, our results suggest that sharing adaptation across experts
can be beneficial compared with maintaining fully separate expert-wise
adapters in MoE fine-tuning.

%% file: camera-ready/08_limitation_acknowledge.tex
\section*{Limitations}
\label{sec:limitation}

Our work focuses on improving the efficiency of PEFT for MoE models. While the method itself does not introduce new model capabilities, it facilitates more efficient adaptation of large language models to downstream tasks. Two limitations are worth noting. First, grouped adapter execution requires routed experts to share the same per-expert input, so it applies to up and gate projections but not to the down projection. Extending the consolidation pipeline to such components is not straightforward and remains for future work. Second, \system{} assumes that expert adapters contain functional redundancy. When a task demands strong expert specialization, excessive consolidation can erase useful distinctions, as shown by the Full Merge ablation in Table~\ref{tab:ace_component_ablationx}, which underperforms expert-wise LoRA (73.57 vs.\ 74.11). \system{} addresses this through similarity-guided grouping rather than uniform merging, but the appropriate grouping granularity (e.g., the resolution parameter \(\gamma\)) may still need tuning when expert specialization is critical. 

\section*{Acknowledgments}

This work was supported by grants from the Institute for Information \&
Communications Technology Planning \& Evaluation (IITP), funded by the
Korea government (MSIT): the Artificial Intelligence Graduate School
Support program (RS-2020-II201336, UNIST), the AI Star Fellowship Program
(RS-2025-25442824, Ulsan National Institute of Science and Technology),
and the Leading Generative AI Human Resources Development program
(IITP-2026-RS-2024-00360227).
This work was also supported by the National Research Foundation of Korea
(NRF) grant funded by the Korea government (MSIT) (RS-2025-00553241).
In addition, this research was supported by the ``Advanced GPU Utilization
Support Program'' funded by the Government of the Republic of Korea
(Ministry of Science and ICT).

We also thank the UAI Lab members, especially Jeonghoon, for their valuable
feedback and support in improving the overall quality of this paper.

%% file: camera-ready/09_appendix.tex
\appendix

\section{Model Configurations}
\label{app:model}
We summarize the key hyperparameters of the MoE models used in our experiments, OLMoE-1B-7B~\citep{muennighoff2025olmoe} and Qwen1.5-MoE-A2.7B~\citep{qwen15moe2024}, { Moonlight-16B-A3B~\citep{liu2025muonscalable}, and Kimi-Linear-48B-A3B~\citep{kimi2025linear},} in Table~\ref{tab:model_configs}. This provides a clear overview of the architectural differences between the models.

\begin{table*}[h]
\centering
\small
\resizebox{\textwidth}{!}{
\begin{tabular}{lcccc}
\toprule
\textbf{Hyperparameter} & \textbf{OLMoE-1B-7B} & \textbf{Qwen1.5-MoE-A2.7B} & \textbf{Moonlight-16B-A3B} & \textbf{Kimi-Linear-48B-A3B} \\
\midrule
Total Parameters & 6.9B & 14.3B & 16B & 48B \\
Active Parameters & 1.3B & 2.7B & 3B & 3B \\
\midrule
Layers ($L$) & 16 & 24 & 27 & 27 \\
Hidden Dimension ($d_{\text{model}}$) & 2048 & 2048 & 2048 & 2304 \\
Number of Experts ($N$) & 64 & 60 + 1 (shared) & 64 + 2 (shared) & 256 + 1 (shared) \\
Top-K Routing ($k$) & 8 & 4 & 6 & 8 \\
Expert FFN Intermediate Size & 1024 & 1408 & 1408 & 1024 \\
\bottomrule
\end{tabular}
}
\caption{A summary of the MoE model configurations used in our experiments.}
\label{tab:model_configs}
\end{table*}

\section{Computing Infrastructure}
\label{appendix:compute_infra}

All experiments were conducted on a workstation running Ubuntu 22.04.5 LTS with CUDA 12.8. Unless otherwise specified, each experiment was conducted using NVIDIA H200 GPUs with 141\,GiB of memory.

For efficiency measurements, wall-clock time is reported as the actual end-to-end elapsed training time, and peak GPU memory is measured using \texttt{torch.cuda.max\_memory\_allocated()}.

\subsection{Configuration for Table~\ref{tab:main_results} and Table~\ref{tab:training_efficiency}}
\subsubsection{Hyperparameters}
\label{app:hyperparams_main}
Unless otherwise specified, all methods were trained under the same optimization setup across the four MoE backbones used in Table~\ref{tab:main_results} and ~\ref{tab:kimi_comparison}: OLMoE-1B-7B-0125, Qwen1.5-MoE-A2.7B, Moonlight-16B-A3B, and Kimi-Linear-48B-A3B. We use the AdamW optimizer with weight decay 0.1 and a constant learning-rate schedule. The maximum sequence length was set to 256, and the effective batch size was 64. For Table~\ref{tab:main_results}, we report results over three random
seeds (0, 1, and 2). For all low-rank adaptation methods, the rank was fixed to $r=2$. We trained all models for 3 epochs, following~\cite{liu2025parameter, hu2023llm}. 
For LoRA methods, the LoRA scaling factor was set to 16 and the LoRA dropout was set to 0.05. 
Following prior
LoRA-based fine-tuning work that uses smaller learning rates for larger
backbones~\citep{zhao2024sapt,xu-etal-2024-bmretriever}, we use $1\times10^{-5}$ for ACE, LoRA, DoRA, and PiSSA on OLMoE
and Qwen, $5\times10^{-6}$ on Moonlight, and $2\times10^{-6}$ on
Kimi-Linear. Our learning rate for OLMoE is also consistent with prior MoE PEFT work~\cite{liu2025parameter}.
AdaLoRA is treated separately, as it required a larger learning rate
than the shared setting used for the other methods~\citep{zhang2023adalora}. We therefore use $7\times10^{-5}$ on OLMoE and Qwen, $4\times10^{-5}$ on Moonlight, and $1\times10^{-5}$ on Kimi-Linear for AdaLoRA.

For \system{}, in addition to the shared training setup above, we set the \(T_{init} = \) 150 and the Louvain resolution parameter $\gamma$ to 0.8. We apply \system{} to the up and gate projections; the down projection retains standard expert-wise LoRA, since its input is already expert-specific and does not admit grouped adapter execution (see Section~\ref{subsec:accel_adapter} for details).


\subsubsection{Baselines}

\begin{itemize}
    \item LoRA~\citep{hu2022lora}: A parameter-efficient fine-tuning method that learns low-rank update matrices while keeping the original model weights frozen.
    \item DoRA~\citep{liu2024dora}: A LoRA-based variant that separates weight updates into directional and magnitude components to improve adaptation quality.
    \item AdaLoRA~\citep{zhang2023adalora}: A dynamic extension of LoRA that redistributes the adaptation rank during training according to the estimated importance of different parameters.
    \item PiSSA~\citep{meng2024pissa}: A low-rank adaptation method that initializes adapters using principal singular components of pretrained weights, leading to more effective and stable optimization.
    \item {PERFT-R~\citep{liu2025parameter}: A MoE-aware PEFT method that introduces separately routed LoRA adapter experts, applying MoE-style routing on top of the adapter modules.}
    \item \revised{EPnG~\citep{lee2026epng}: A MoE-aware PEFT method that reallocates LoRA capacity across experts according to their routing importance, assigning larger adaptation capacity to more important experts.}

\end{itemize}

\subsubsection{Notes on MoE-Specific Baselines}
We provide additional context on how MoE-specific PEFT methods are positioned in our comparison.

\paragraph{PERFT-R.}
PERFT-R~\citep{liu2025parameter} is a MoE-aware PEFT method that introduces MoE-style routing over LoRA adapters through separately routed adapter experts.
We include PERFT-R as a MoE-specific PEFT baseline because it explicitly incorporates MoE structure into adapter tuning, unlike generic PEFT methods adapted from dense Transformers.
We note that our reported numbers may differ from those in the original PERFT-R paper, as we evaluate all methods under a unified setup that matches the expert-wise LoRA configuration used for the other baselines in this work (e.g., adapted projections, parameter budget, and optimization hyperparameters), rather than the configuration adopted in the original PERFT-R evaluation.
This setup is chosen to ensure a fair comparison across all PEFT baselines under matched conditions.

\paragraph{ESFT.}
ESFT~\citep{wang2024esft} selects task-relevant experts and directly fine-tunes their parameters. It is therefore a selective expert tuning method rather than a LoRA-style adapter-based PEFT method. Since its trainable parameter count depends on the number of selected experts and tuned layers, ESFT is difficult to match to the adapter-based budget used by ACE and the LoRA baselines. For a fair comparison, we restrict the main quantitative evaluation to methods under the same LoRA-based PEFT setting, and discuss ESFT as related MoE fine-tuning work in Section \ref{sec:related_work}.

\subsubsection{Seed-wise Consistency}
\label{app:seed_consistency}

To further characterize run-to-run variability in the main results,
we compare \system{} with the strongest fully evaluated non-\system{}
baseline for each backbone in Table~\ref{tab:main_results}.

\begin{table}[t]
\centering
\footnotesize
\setlength{\tabcolsep}{4pt}
\renewcommand{\arraystretch}{1.08}
\begin{tabular}{llccc}
\toprule
\textbf{Backbone} & \textbf{Method}
& \textbf{Seed 0} & \textbf{Seed 1} & \textbf{Seed 2} \\
\midrule
OLMoE & PiSSA     & 74.58 & 74.90 & 74.78 \\
      & \system{} & \textbf{75.70} & \textbf{75.84} & \textbf{75.30} \\
\midrule
Qwen  & PiSSA     & 80.11 & 79.96 & 79.89 \\
      & \system{} & \textbf{80.32} & \textbf{80.42} & \textbf{80.07} \\
\midrule
Moonlight & EPnG      & 79.89 & 80.13 & 80.36 \\
          & \system{} & \textbf{80.98} & \textbf{81.35} & \textbf{81.19} \\
\bottomrule
\end{tabular}
\caption{Seed-wise average accuracy against the strongest fully evaluated
non-\system{} baseline for each backbone.}
\label{tab:seed_consistency}
\end{table}

Across all three random seeds, \system{} consistently achieves higher
average accuracy than the corresponding strongest baseline on each
backbone. In total, \system{} is higher in all nine matched
backbone--seed comparisons. These results indicate that the improvements
reported in Table~\ref{tab:main_results} are consistent across random
seeds.

\subsection{Configuration for Table~\ref{tab:math_code_legal_results}}
\label{app:math_code_details}
For mathematical reasoning and code generation, we use the same optimization setup as in Section~\ref{app:hyperparams_main}, except that the maximum sequence length was set to 1024, the effective batch size was 16, the rank was set to 8, and all models were trained for 1 epoch.
For code datasets, we report pass@10 using a sampling temperature of 0.7; for math datasets, we report exact-match accuracy.
For legal understanding, we follow the optimization setup of Section~\ref{app:hyperparams_main} with rank 8 and report accuracy.

\begin{table*}[h]

\centering
\small
\resizebox{0.88\textwidth}{!}{
\begin{tabular}{llccccccccc}
\toprule
Method & $T_\text{init}$ & HS & WG & ARC-c & ARC-e & BoolQ & OBQA & PIQA & SIQA & Avg. \\
\midrule
LoRA      & --  & 90.82 & 70.08 & 69.42 & 85.61 & 66.71 & 74.60 & 83.10 & 52.51 & 74.11 \\
\system{} & 50  & 92.14 & 72.45 & 71.33 & 86.62 & 69.36 & 77.40 & 82.37 & 53.38 & 75.63 \\
\system{} & 150 & 92.27 & 73.32 & 70.82 & 86.45 & 69.14 & 76.60 & 84.06 & 52.97 & \textbf{75.70} \\
\system{} & 500 & 92.52 & 70.88 & 69.62 & 86.41 & 68.78 & 76.40 & 83.73 & 53.53 & 75.23 \\
\bottomrule
\end{tabular}
}
\caption{{Effect of warm-up duration on \system{}. We vary $T_\text{init}$ while keeping all other settings fixed, and include LoRA as a baseline. \system{} shows limited sensitivity to the exact warm-up duration and outperforms LoRA at every $T_\text{init}$, with $T_\text{init}=150$ giving the best average performance in this sweep. Results are from one random seed.}}
\label{tab:tinit_sensitivity}
\end{table*}

{
\ifbluemode\color{accentblue}\fi
\subsection{Dataset Statistics}
\label{app:data_statistics}

We use four training datasets comprising 764,455 source examples:
Commonsense170K (170,420), MetaMathQA (395,000),
CodeFeedback-Filtered-Instruction (156,526), and CaseHOLD (42,509).
For the commonsense reasoning results in Table~\ref{tab:main_results}, following~\citealp{hu2023llm},
Commonsense170K combines the training splits of eight commonsense tasks.
After tokenization and sequence packing, each training set is randomly
split into 98\% training and 2\% validation subsets.

For evaluation, we use 30,403 examples across 12 benchmarks:
HellaSwag (10,042), WinoGrande (1,267), ARC-Challenge (1,172),
ARC-Easy (2,376), BoolQ (3,270), OpenBookQA (500), PIQA (1,838),
Social IQA (1,954), GSM8K (1,319), MATH  (1,187),
HumanEval (164), and CaseHOLD (5,314).
}

\subsection{Configuration for Table~\ref{tab:ace_component_ablationx}}
\label{app:ablation_details}
The ablation study in Table~\ref{tab:ace_component_ablationx} is conducted on OLMoE-1B-7B-0125 using the same training setup and hyperparameters as Table~\ref{tab:main_results} (Appendix~\ref{app:hyperparams_main}).
All compared methods share the same optimization configuration. The trainable parameter budget is matched for LoRA, Random, Full Merge, w/o Grp. Exec., and \system{}, while w/o Rank Cons. is reported separately because disabling rank consolidation changes the number of trainable adapter parameters.

\subsection{Licenses and Terms of Use}
\label{app:licenses}
All model backbones and datasets used in this work are publicly available,
and we cite their original sources throughout the paper.
OLMoE-1B-7B-0125 is released under Apache-2.0,
Qwen1.5-MoE-A2.7B under the Tongyi Qianwen License,
and Moonlight-16B-A3B and Kimi-Linear-48B-A3B under the MIT License.
For the datasets, we use only their publicly released versions for
research evaluation and follow the applicable licenses and terms of use
specified by their original providers. Some datasets do not provide
unambiguous standalone license information in their original releases;
for these resources, we follow the usage conditions provided by the
corresponding authors or benchmark distributions.

\section{Sensitivity and Robustness to Hyperparameters}
\label{app:hyperparameter_sensitivity}
\subsection{Stability Analysis for Fixed Grouping and Warm-up Duration}
\label{app:merge_step_stability}

\paragraph{Local grouping stability.} To justify using a fixed grouping after warm-up, we examine the trajectory of expert grouping over training across the model. Every 15 optimization steps, we compute the expert similarity matrices and apply the same grouping procedure as in the main method, yielding a grouping structure at step $t$.

For each step, we convert the partition into the set of expert pairs assigned to the same group,
\[
\mathcal{S}^{(t)}=\{(i,j)\mid 1\le i<j\le E,\ \Pi^{(t)}(i)=\Pi^{(t)}(j)\}.
\]
We then compare the grouping at step \(t\) with the grouping from the previous measurement step \(t-\Delta\) using Jaccard similarity:
\[
J_{\mathrm{local}}(t)=
\frac{|\mathcal{S}^{(t)}\cap \mathcal{S}^{(t-\Delta)}|}
     {|\mathcal{S}^{(t)}\cup \mathcal{S}^{(t-\Delta)}|}.
\]
Here, \(\Delta=15\) corresponds to the interval between two consecutive grouping measurements.

\begin{figure}[h]
    \centering
    \includegraphics[width=\linewidth]{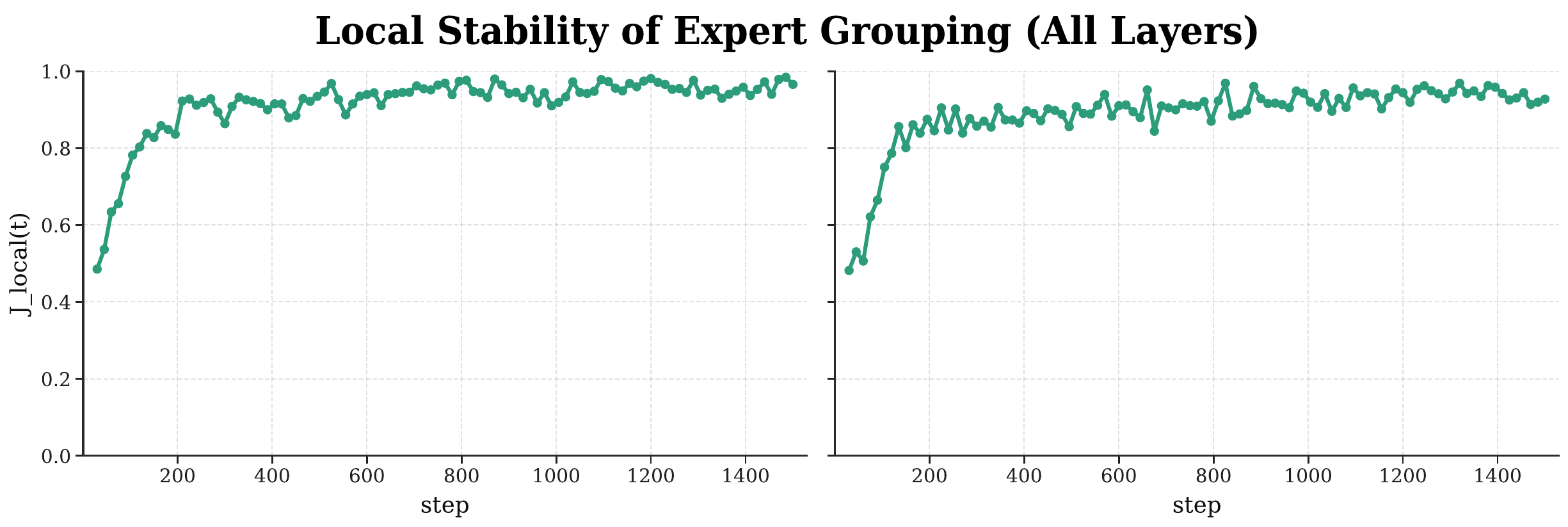}
    \caption{Local stability of expert grouping over training in the all-layer view. Every 15 steps, we recompute the grouping and compare it with the grouping from the previous measurement step using Jaccard similarity. The left and right panels correspond to training on the HellaSwag and WinoGrande datasets, respectively. In both settings, the local similarity increases quickly and then plateaus, indicating that the grouping structure has stabilized.}
    \label{fig:merge_step_stability}
\end{figure}

Figure~\ref{fig:merge_step_stability} shows that the grouping changes rapidly early in training and then enters a stable regime, where consecutive grouping estimates become highly similar.
This supports our design choice of using a single grouping after warm-up, since the grouping structure does not continue to change substantially once it enters the stable regime.
We choose $T_\text{init}=150$ as a conservative default: it lies in the stable regime while still leaving most of training for shared-adapter optimization.

\paragraph{Effect on routing statistics.}
ACE keeps the pretrained router frozen and applies the fixed group
assignment only after expert selection. To examine whether fixed grouping
alters routing behavior, we compare aggregate routing statistics between
ACE and expert-wise LoRA throughout training.

\begin{table}[t]
\centering
\footnotesize
\setlength{\tabcolsep}{4pt}
\renewcommand{\arraystretch}{1.08}
\begin{tabular}{lcc}
\toprule
\textbf{Step}
& \textbf{Load Entropy} $\uparrow$
& \textbf{Aux. Loss} $\downarrow$ \\
& \textbf{LoRA / ACE}
& \textbf{LoRA / ACE} \\
\midrule
500   & 0.9613 / 0.9605 & 1.1168 / 1.1214 \\
1000  & 0.9614 / 0.9616 & 1.1161 / 1.1162 \\
Final & 0.9619 / 0.9620 & 1.1142 / 1.1133 \\
\bottomrule
\end{tabular}
\caption{
Routing statistics for LoRA and ACE during training.
}
\label{tab:routing_stats}
\end{table}

Across training, the maximum absolute differences between ACE and LoRA
are only 0.0008 in load entropy and 0.0046 in auxiliary loss.
These results indicate that fixed group assignments have little effect
on the measured aggregate routing statistics in this setting.

{
\ifbluemode\color{accentblue}\fi
\paragraph{CKA consistency with later training.}
To assess whether the similarity structure identified during warm-up
remains representative later in training, we compare intermediate CKA
matrices with those at the final checkpoint. For this post-hoc diagnostic,
we recompute centered linear CKA at each checkpoint using the same fixed
10,000-token held-out validation probe, which is distinct from the training
examples used for consolidation (Appendix~\ref{app:louvain}). This allows us to compare
checkpoints under a common, unseen input set. For each of the 16 MoE layers
and two projection types (\texttt{gate\_proj} and \texttt{up\_proj}), we
vectorize the strict upper triangle of the $64\times64$ CKA matrix,
corresponding to 2,016 unique expert pairs, and compute its Pearson and
Spearman correlations with the corresponding vector at the final checkpoint
(step 2,805). We report the unweighted mean across the resulting 32
layer--projection combinations. The Pearson/Spearman correlations increase
from 0.577/0.587 at step 150 to 0.771/0.772 at step 500 and 0.901/0.896
at step 1000, indicating that the structure identified at our default
consolidation step already shows moderate alignment with the final structure
and becomes progressively more consistent during training. 

However, stronger CKA agreement does not necessarily lead to better downstream performance:
consolidation at step 500 yields a slightly lower average accuracy than at
step 150 (75.23 vs.\ 75.70; Table~\ref{tab:tinit_sensitivity}). These results
suggest that $T_{\mathrm{init}}$ should balance structural reliability with
the remaining optimization time after consolidation, rather than simply
maximizing agreement with the final checkpoint.

\paragraph{Sensitivity to warm-up duration.} Table~\ref{tab:tinit_sensitivity} shows that performance is stable across $T_\text{init}$, with a maximum gap of only 0.47 points between configurations.
This robustness stems from an inherent trade-off: a larger $T_\text{init}$ yields more stable grouping estimates, while a smaller $T_\text{init}$ lets the shared adapters benefit from group-level updates for longer.
These two competing effects offset each other, making \system{} relatively insensitive to the exact choice of $T_\text{init}$.
Notably, \system{} outperforms LoRA (74.11) across all $T_\text{init}$ settings, confirming robustness to this hyperparameter.
We adopt $T_\text{init}=150$ as the default, which yields the best average score in this sweep.

}

\paragraph{Practical guidance.}
The warm-up duration \(T_{\mathrm{init}}\) controls a trade-off between grouping reliability and remaining optimization time after consolidation. 
Consolidating too early can use unstable adapter representations, whereas consolidating too late leaves limited time for shared adapters to benefit from group-level updates. 
Our sensitivity results show that \system{} is robust within a reasonable range, so \(T_{\mathrm{init}}\) does not need to be finely tuned. 
In practice, we recommend consolidating after the steepest initial loss decrease, when training begins to enter a more gradual descent, but well before the majority of training is complete.


\subsection{Sensitivity to the Louvain Resolution Parameter}
\label{app:gamma}

We analyze how the Louvain resolution parameter $\gamma$ affects performance and efficiency on OLMoE-1B-7B-0125 across the eight commonsense reasoning benchmarks of Section~\ref{sub:commonsense}.
A larger $\gamma$ produces more, smaller groups (approaching one adapter per expert), while a smaller $\gamma$ produces fewer, larger groups (approaching full consolidation).

\begin{table}[h]

\centering
\small
\resizebox{\columnwidth}{!}{
\setlength{\tabcolsep}{6pt}
\renewcommand{\arraystretch}{1.15}
\begin{tabular}{lccc}
\toprule
Method & Group Count & Avg.$\uparrow$ & Eval Time (s)$\downarrow$ \\
\midrule
LoRA                                & --  & 74.11 & 20.82 \\
\midrule
\system{} ($\gamma=3.2$)            & 242 & 74.59 & 19.76 \\
\system{} ($\gamma=1.6$)            & 215 & 75.04 & 17.67 \\
\system{} ($\gamma=1.2$)            & 141 & 75.21 & 17.24 \\
\system{} ($\gamma=0.8$, default)   & 97  & 75.61 & 17.22 \\
\system{} ($\gamma=0.4$)            & 67  & 75.75 & 16.97 \\
\system{} ($\gamma=0.1$)            & 46  & 75.82 & 16.85 \\
\midrule
\system{} (full merge, $\gamma\!\to\!0$) & --  & 73.57 & 16.41 \\
\bottomrule
\end{tabular}
}
\caption{
{Effect of the Louvain resolution parameter $\gamma$ on OLMoE. Group Count is the total number of consolidated adapter groups across all MoE layers. Avg.\ is the eight-benchmark commonsense reasoning average. Eval time is the wall-clock duration of one full evaluation pass on a single H200.}}
\label{tab:gamma_sensitivity}
\end{table}

Table~\ref{tab:gamma_sensitivity} shows three consistent trends.
First, \system{} \emph{robustly} outperforms LoRA across all evaluated group counts (46-242), with average accuracy varying by only 1.23 points, so $\gamma$ does not require careful tuning.
Second, over-aggressive consolidation hurts performance: fully merging all experts ($\gamma\!\to\!0$) drops below LoRA (73.57 vs.\ 74.11), indicating that some degree of expert specialization must be preserved.
Third, evaluation time decreases monotonically as $\gamma$ shrinks (fewer, larger groups), achieving up to a 19.1\% reduction over LoRA at $\gamma=0.1$.
We adopt $\gamma=0.8$ as the default because it offers a robust accuracy-efficiency trade-off (+1.50 over LoRA, $-17.3\%$ eval time) and shows the most consistent behavior across models and datasets in our experiments.

At the default $\gamma=0.8$, \system{} forms 97 multi-expert
consolidated groups together with 42 singleton experts that remain
unmerged. Among the consolidated groups, the median group size is 23
experts and the largest contains 43 experts. This places \system{}
between expert-wise LoRA and uniform full consolidation.

\subsection{Necessity of Centering in CKA Calculation}
\label{app:cka_centering}

Standard CKA mean-centers representation matrices over the probe set before computing similarity~\citep{kornblith2019similarity}.
Since expert activations in ACE already reside in a shared feature space, centering is not strictly required from a coordinate-alignment perspective.
We use centered CKA simply for consistency with the standard formulation.
To verify that this choice does not affect our results, we compare centered and uncentered CKA at the grouping step on OLMoE.

\begin{table}[h]

\centering
\resizebox{\columnwidth}{!}{
\setlength{\tabcolsep}{6pt}
\renewcommand{\arraystretch}{1.12}
\begin{tabular}{lcc}
\toprule
\textbf{Metric} & \textbf{Centered CKA} & \textbf{Uncentered CKA} \\
\midrule
Raw MSE Gap $\uparrow$                 & -2.417e-6 & 	-3.911e-6 \\
Louvain Q $\uparrow$                   & 0.395                   & 0.374                   \\
Downstream Average Accuracy $\uparrow$ & 75.70                   & 75.44                   \\
\bottomrule
\end{tabular}
}
\caption{Comparison between centered and uncentered CKA on OLMoE commonsense reasoning. Both variants use the same ACE pipeline and differ only in whether the representation matrices are mean-centered before computing CKA similarity. Results are from one random seed.}
\label{tab:cka_centering}
\end{table}

The two variants yield broadly comparable grouping quality and downstream performance, confirming that the centering choice does not critically affect \system{}.

{
\ifbluemode\color{accentblue}\fi

\subsection{Sensitivity to Probe-Set Size}
\label{app:probe_size}

To examine whether the expert grouping discovered by \system{} is
sensitive to the probe-set size, we compare group assignments obtained
with smaller probe sets against the default 10k-token reference in the
OLMoE commonsense setting.

\begin{table}[t]
\centering
\ifbluemode\color{accentblue}\fi
\footnotesize
\setlength{\tabcolsep}{4pt}
\renewcommand{\arraystretch}{1.08}
\begin{tabular}{lccc}
\toprule
\textbf{Probe Tokens}
& \textbf{Jaccard}
& \textbf{Pair-F1}
& \textbf{ARI} \\
\midrule
512    & 0.923 & 0.953 & 0.922 \\
1,000  & 0.946 & 0.969 & 0.948 \\
2,000  & 0.958 & 0.976 & 0.960 \\
5,000  & 0.989 & 0.994 & 0.991 \\
10,000 & 1.000 & 1.000 & 1.000 \\
\bottomrule
\end{tabular}
\caption{Similarity of group assignments obtained with different
probe-set sizes to the default 10k-token reference on OLMoE.}
\label{tab:probe_size}
\end{table}

Even 512 probe tokens recover a highly similar grouping
(ARI $=0.922$), while 5k tokens produce nearly identical assignments
to the 10k-token reference. These results indicate that \system{} is
robust to substantial reductions in probe-set size.
}

\section{Rank Variation and Optimization Dynamics}
\label{app:rank_variation}

\begin{figure}[h]
    \centering
    \begin{subfigure}[t]{\columnwidth}
        \centering
        \includegraphics[width=0.9\columnwidth]{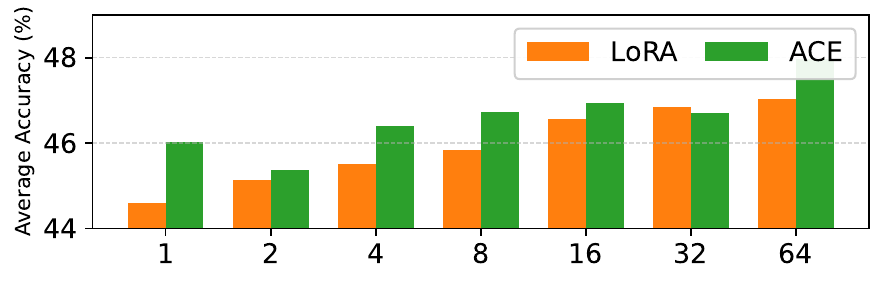}
        \caption{{Average accuracy on GSM8K and MATH across ranks.}}
        \label{fig:rank_robustness_avg}
    \end{subfigure}
    
    \vspace{0.2em}
    
    \begin{subfigure}[t]{\columnwidth}
        \centering
        \includegraphics[width=0.9\columnwidth]{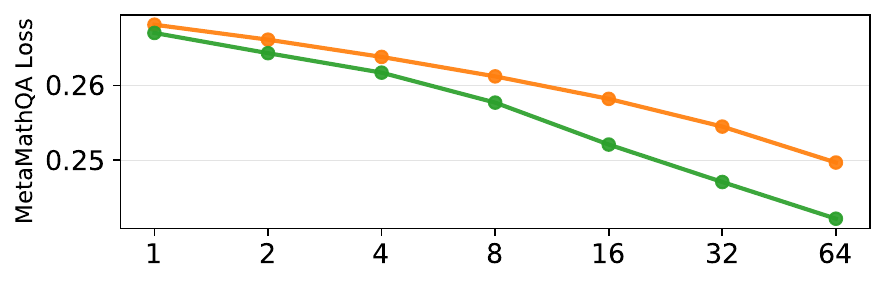}
        \caption{{Final training loss across ranks.}}
        \label{fig:rank_sweep_final_loss}
    \end{subfigure}
    
    \caption{
    {
    Rank sensitivity analysis of ACE and LoRA. 
    ACE maintains stronger average performance across different adapter ranks and shows lower final training loss, indicating robustness to the choice of adaptation capacity.
    }}
    \label{fig:rank_sensitivity}
\end{figure}

To examine whether the gains of ACE depend on the adapter rank used in the main experiments, we compare ACE and LoRA across $r \in \{1,2,4,8,16,32,64\}$. Figure \ref{fig:rank_sensitivity}(a) reports the average accuracy over GSM8K and MATH, and Figure \ref{fig:rank_sensitivity}(b) reports the final training loss at each rank.

As shown in Figure \ref{fig:rank_sensitivity}(a), ACE maintains higher average accuracy than LoRA across most rank settings. Importantly, this trend holds not only at the main experimental setting, but also when the adapter capacity is made smaller or larger. This indicates that the improvement of ACE is not tied to a particular rank choice, but is consistently observed across different adaptation-capacity regimes.

Figure \ref{fig:rank_sensitivity}(b) provides optimization-side evidence for the same trend. ACE obtains lower final training loss than LoRA across ranks, suggesting that the accuracy improvement is accompanied by a more favorable training objective rather than appearing only in downstream evaluation. Taken together, the rank-wise accuracy and final-loss results show that ACE remains robust to the choice of adapter rank and continues to provide stable gains when the available adaptation capacity changes.

\section{Analysis of Group-Anchor Strategies}
\label{sub:anchor}

To minimize functional discontinuity during consolidation, \system{} initializes each shared adapter from a similarity-central anchor within the group while assigning the remaining rank dimensions using standard LoRA initialization. We further compare three strategies for constructing this group anchor. Avg.\ $\Delta W$ forms a representative update by averaging the expert-specific full updates within a group and then refitting the result to the low-rank LoRA form via SVD. Max $\Delta W$ selects the expert with the largest update magnitude as the anchor. CKA-center selects the expert that is most central with respect to the within-group CKA similarities, thereby matching the similarity-based initialization principle used in our methodology.

\begin{table}[h]
\centering
\small
\resizebox{\columnwidth}{!}{
\begin{tabular}{lcccc}
\toprule
Method & HellaSwag & WinoGrande & OBQA & Average \\
\midrule
Avg.\ $\Delta W$ & 92.52 & 67.17 & 76.60 & 78.76 \\
Max $\Delta W$ & 92.46 & 68.27 & 76.40 & 79.04 \\
CKA-center & 92.27 & 68.98 & 76.60 & 79.28 \\
\bottomrule
\end{tabular}
}
\caption{Analysis of group-anchor strategies for shared-adapter initialization.}
\label{tab:anchor_selection}
\end{table}

As shown in Table~\ref{tab:anchor_selection}, the three strategies yield broadly comparable results, but CKA-center provides the strongest overall performance. It is also the most efficient choice. Avg.\ $\Delta W$ introduces the highest overhead because it requires constructing an averaged full update and performing an additional SVD refitting step. Max $\Delta W$ also incurs extra computation due to the additional update-magnitude scoring required for anchor selection. In contrast, CKA-center directly reuses the similarity statistics already computed during grouping and therefore adds essentially no extra cost. Since it is both the most effective and the most efficient option, we adopt CKA-center as the default group-anchor strategy.

\begin{table*}[t]
\centering
\small
\setlength{\tabcolsep}{5pt}
\renewcommand{\arraystretch}{1.15}

\resizebox{\textwidth}{!}{
\begin{tabular}{l l c c c c c c c c c}
\toprule
\textbf{Model} & \textbf{Method} & \textbf{\# Params (\%)} 
& \textbf{HellaSwag} & \textbf{WinoGrande} 
& \textbf{ARC-c} & \textbf{ARC-e} 
& \textbf{BoolQ} & \textbf{OBQA} & \textbf{PIQA}
& \textbf{Avg.} \\
\midrule

\multirow{4}{*}{\textbf{OLMoE}}
& LoRA       & 0.27 & 87.58 & 64.09 & 53.92 & 67.85 & 73.55 & 64.40 & 70.13 & 68.79 \\
& AdaLoRA     & 0.54 $\rightarrow$ 0.27 & 86.06 & 62.27 & \textbf{55.63} & 73.99 & 71.83 & 62.60 & 64.64 & 68.15 \\
& DoRA         & 0.33 & 87.41 & 63.06 & 52.13 & 69.99 & 75.90 & 60.80 & 74.97 & 69.18 \\
& \cellcolor{ourscolor}\textbf{ACE (Ours)} 
& \cellcolor{ourscolor}0.27 
& \cellcolor{ourscolor}\textbf{89.08} 
& \cellcolor{ourscolor}\textbf{71.11} 
& \cellcolor{ourscolor}55.46 
& \cellcolor{ourscolor}\textbf{75.29} 
& \cellcolor{ourscolor}\textbf{78.90} 
& \cellcolor{ourscolor}\textbf{64.60} 
& \cellcolor{ourscolor}\textbf{78.51} 
& \cellcolor{ourscolor}\textbf{73.28} \\
\midrule

\multirow{4}{*}{\textbf{Qwen}}
& LoRA        & 0.21 & 91.82 & 77.03 & 73.98 & 87.84 & 83.70 & 78.60 & 84.77 & 82.53 \\
& AdaLoRA      & 0.42 $\rightarrow$ 0.21 & 90.19 & 65.90 & \textbf{75.26} & \textbf{88.30} & 73.24 & 60.60 & 82.48 & 76.57 \\
& DoRA         & 0.26 & 91.67 & 75.37 & 72.10 & 87.54 & 83.88 & 77.20 & 84.11 & 81.70 \\
& \cellcolor{ourscolor}\textbf{ACE (Ours)} 
& \cellcolor{ourscolor}0.21 
& \cellcolor{ourscolor}\textbf{92.61} 
& \cellcolor{ourscolor}\textbf{77.19} 
& \cellcolor{ourscolor}74.40
& \cellcolor{ourscolor}88.01 
& \cellcolor{ourscolor}\textbf{84.16} 
& \cellcolor{ourscolor}\textbf{79.80} 
& \cellcolor{ourscolor}\textbf{84.98} 
& \cellcolor{ourscolor}\textbf{83.02} \\
\bottomrule
\end{tabular}
}
\caption{
Accuracy (\%) on commonsense reasoning benchmarks. 
All methods use the same adapter rank on MoE experts, resulting in nearly 
identical trainable parameter budgets.
}
\label{tab:per_dataset_result}
\end{table*}

\section{Details for Figure~\ref{fig:similarity_matrix}}
\label{app:cka}
Figure~\ref{fig:similarity_matrix} presents the pairwise CKA similarity matrix of OLMoE-1B-7B-0125, computed using the gate projection outputs from the 0-th, 4-th and 14-th layers. The similarity is measured for all pairs of experts, resulting in a full pairwise similarity matrix. The CKA scores used for the visualization are computed from the HellaSwag dataset. Values closer to 1 indicate higher similarity, while values closer to 0 indicate lower similarity.

\section{Results with Dataset-Specific Training}
\label{app:split_dataset}

In addition to the main experiments, we evaluate our method in a dataset-specific training setting, where a separate model is trained for each dataset, to analyze its behavior on individual benchmarks. We use two MoE backbones, OLMoE-1B-7B-0125 and Qwen1.5-MoE-A2.7B, and report results on HellaSwag, WinoGrande, ARC-Challenge, ARC-Easy, BoolQ, OpenBookQA, and PIQA.

We compare our approach with LoRA, AdaLoRA, and DoRA. For a fair comparison, all methods apply adapters to the expert modules and use the same rank configuration within each setting, so that the number of trainable parameters remains nearly identical across methods. Unless otherwise noted, we use the same training setup and hyperparameters as in Appendix~\ref{app:hyperparams_main}, except that each model is trained for 1{,}500 steps.

As shown in Table~\ref{tab:per_dataset_result}, \system{} achieves stronger performance than the competing methods on most benchmarks and consistently outperforms the baselines across several datasets.

{
\ifbluemode\color{accentblue}\fi
\section{Free-Form Generation on XSum}
\label{app:xsum}

To evaluate generation quality beyond accuracy and pass@10, we evaluate
PEFT methods on the XSum~\cite{narayan2018don} summarization benchmark using OLMoE.
All methods are compared under matched trainable-parameter budgets.

\begin{table}[t]
\ifbluemode\color{accentblue}\fi
\centering
\footnotesize
\setlength{\tabcolsep}{5pt}
\renewcommand{\arraystretch}{1.08}
\begin{tabular}{lccc}
\toprule
\textbf{Method}
& \textbf{ROUGE-1} $\uparrow$
& \textbf{ROUGE-2} $\uparrow$
& \textbf{ROUGE-L} $\uparrow$ \\
\midrule
LoRA    & 32.98 & 13.77 & 26.27 \\
DoRA    & 33.16 & 13.90 & 26.45 \\
AdaLoRA & 30.01 & 11.82 & 24.17 \\
PiSSA   & 30.48 & 12.26 & 24.72 \\
PERFT-R & 33.57 & 14.31 & 26.90 \\
\system{} & \textbf{33.82} & \textbf{14.37} & \textbf{27.01} \\
\bottomrule
\end{tabular}
\caption{
XSum summarization results on OLMoE under matched
trainable-parameter budgets.
}
\label{tab:xsum}
\end{table}

\system{} achieves the highest score on all three ROUGE metrics, indicating that its
gains extend to free-form text generation.

}

\section{Additional Functional Redundancy Analyses}
\label{app:functional_redundancy}

\subsection{Expert Swap on Up Projection}
\label{app:swapping_up_proj}

\begin{figure}[h!]
    \centering
    \includegraphics[width=1\linewidth]{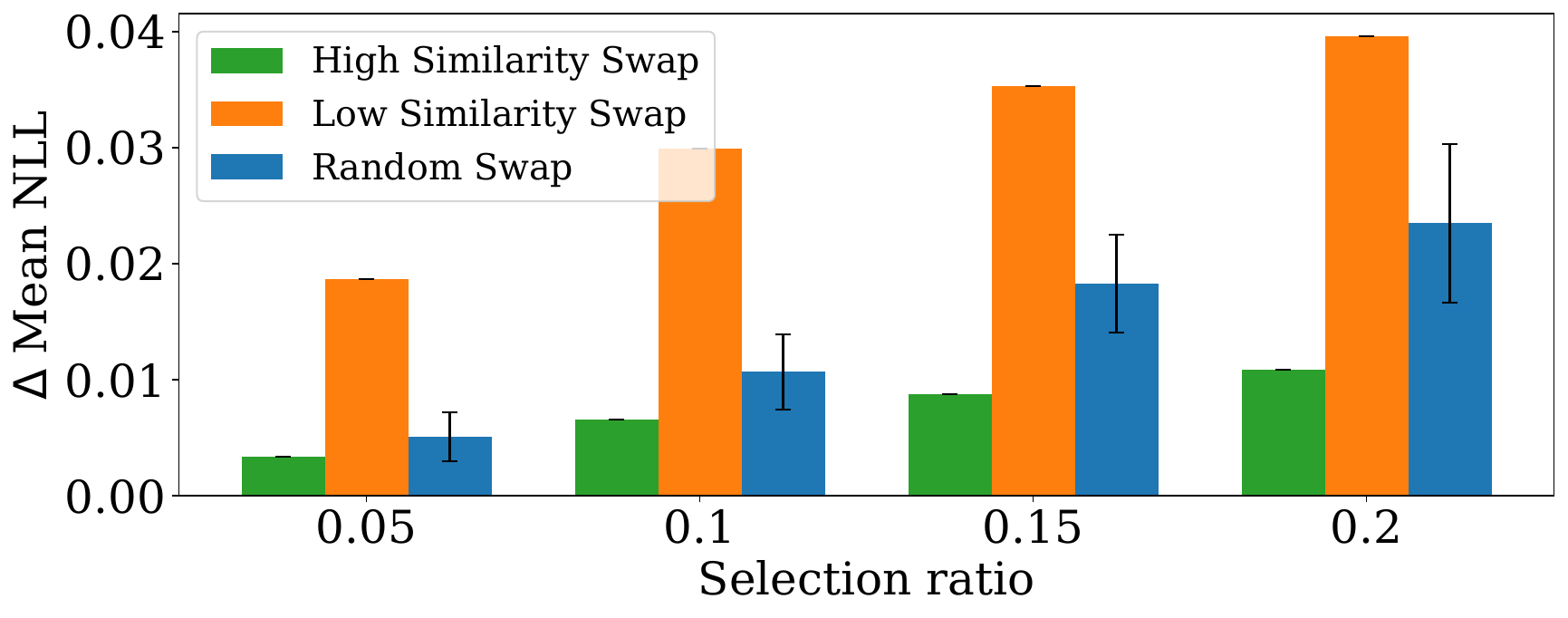}
    \caption{Effect of adapter swapping based on expert similarity applied on up projection. Replacing adapters with high similarity experts causes only minor increases in NLL, while dissimilar or random experts lead to substantially larger degradation.}
    \label{fig:expert_swap_up}
\end{figure}

We conduct this analysis on OLMoE-1B-7B-0125 with $\Delta\mathrm{NLL}$ evaluated on HellaSwag. Figure~\ref{fig:expert_swap_up} reveals a strong dependence on LoRA similarity. Replacing LoRA with highly similar ones results in only minor increases in Negative Log Likelihood (NLL) across all selection ratios. For instance, when \(\rho=0.05\), the NLL rises by approximately \(0.0034\) relative to the baseline. In contrast, swapping with low-similarity adapters leads to substantially larger degradation, increasing the NLL by \(0.0187\), while random swaps produce a moderate increase of about \(0.0051\).

This trend persists as the replacement ratio increases. At \(\rho=0.20\), high-similarity swaps increase the NLL by only \(0.0109\), whereas low-similarity swaps cause a much larger rise of \(0.0396\). Random swaps again fall between these two extremes, increasing the NLL by \(0.0235\).

Taken together, these results indicate that CKA similarity reliably reflects functional interchangeability between expert adapters. Experts with highly aligned adapter responses can therefore substitute for one another with minimal performance degradation, supporting the design principle of \system{} that adaptation capacity can be consolidated across redundant experts.

\subsection{Functional Redundancy across Models and Tasks}
\label{app:cka_cross_task}

Across commonsense reasoning, the CKA gaps are 0.299, 0.298, 0.438,
and 0.344 for OLMoE, Qwen, Moonlight, and Kimi, respectively.
We additionally observe positive gaps on mathematical reasoning
(0.259/0.215), code generation (0.251/0.208), and legal understanding
(0.323/0.286) for OLMoE/Qwen, respectively.

To verify that this separation is not specific to the data used for
group construction, we construct groups on one data split and evaluate
their CKA separation on an independent split. In the OLMoE commonsense
setting, the discovered groups retain a CKA gap of 0.293 on the held-out
split, whereas random groupings yield a near-zero gap of
$-0.0006 \pm 0.0034$.

{

\section{Extended Backbone Comparison Including Kimi}
\label{app:kimi_comparison}

To examine whether \system{}'s gains over expert-wise LoRA generalize to a larger MoE backbone, we extend the comparison to Kimi-Linear-48B-A3B.
Table~\ref{tab:kimi_comparison} compares \system{} with LoRA across all four MoE backbones evaluated in this work.
For each backbone we report the eight-benchmark average and per-task accuracies, with results for OLMoE, Qwen1.5-MoE, and Moonlight from Table~\ref{tab:main_results}.
\system{} consistently improves over LoRA on every backbone, including
Kimi (84.18 vs.\ 83.66), indicating that the gains of adapter
consolidation extend to a larger backbone with substantially more experts.
}

\begin{table*}[t]

\centering
\renewcommand{\arraystretch}{1.15}
\resizebox{\textwidth}{!}{
\begin{tabular}{l l c c c c c c c c c c}
\toprule
\textbf{Model} & \textbf{Method} & \textbf{\# Params (\%)$\downarrow$}
& \textbf{HS}$\uparrow$ & \textbf{WG}$\uparrow$
& \textbf{ARC-c}$\uparrow$ & \textbf{ARC-e}$\uparrow$
& \textbf{BoolQ}$\uparrow$ & \textbf{OBQA}$\uparrow$ & \textbf{PIQA}$\uparrow$
& \textbf{SIQA}$\uparrow$ & \textbf{Avg.}$\uparrow$ \\
\midrule

\multirow{2}{*}{\textbf{OLMoE} (1B/7B)}
& LoRA        & 0.27 & 90.82 & 70.08 & 69.42 & 85.61 & 66.71 & 74.60 & 83.10 & 52.51 & 74.11 \\
& \cellcolor{ourscolor}\textbf{\system{}}
& \cellcolor{ourscolor}0.27
& \cellcolor{ourscolor}\textbf{92.24} & \cellcolor{ourscolor}\textbf{72.85}
& \cellcolor{ourscolor}\textbf{70.88} & \cellcolor{ourscolor}\textbf{86.21}
& \cellcolor{ourscolor}\textbf{68.04} & \cellcolor{ourscolor}\textbf{77.33}
& \cellcolor{ourscolor}\textbf{83.93} & \cellcolor{ourscolor}\textbf{53.41}
& \cellcolor{ourscolor}\textbf{75.61} \\
\midrule

\multirow{2}{*}{\textbf{Qwen} (3B/14B)}
& LoRA        & 0.21 & 93.23 & 77.66 & 79.44 & 90.63 & 70.84 & 82.67 & \textbf{86.80} & 56.72 & 79.75 \\
& \cellcolor{ourscolor}\textbf{\system{}}
& \cellcolor{ourscolor}0.21
& \cellcolor{ourscolor}\textbf{93.81} & \cellcolor{ourscolor}\textbf{79.19}
& \cellcolor{ourscolor}\textbf{79.50} & \cellcolor{ourscolor}\textbf{90.66}
& \cellcolor{ourscolor}\textbf{71.49} & \cellcolor{ourscolor}\textbf{83.87}
& \cellcolor{ourscolor}86.67 & \cellcolor{ourscolor}\textbf{56.99}
& \cellcolor{ourscolor}\textbf{80.27} \\
\midrule

\multirow{2}{*}{\textbf{Moonlight} (3B/16B)}
& LoRA        & 0.22 & 92.23 & 74.59 & \textbf{84.67} & 94.28 & 67.17 & 82.00 & 86.92 & 52.47 & 79.29 \\
& \cellcolor{ourscolor}\textbf{\system{}}
& \cellcolor{ourscolor}0.22
& \cellcolor{ourscolor}\textbf{93.49} & \cellcolor{ourscolor}\textbf{79.95}
& \cellcolor{ourscolor}84.30 & \cellcolor{ourscolor}\textbf{94.75}
& \cellcolor{ourscolor}\textbf{69.31} & \cellcolor{ourscolor}\textbf{84.93}
& \cellcolor{ourscolor}\textbf{87.71} & \cellcolor{ourscolor}\textbf{54.93}
& \cellcolor{ourscolor}\textbf{81.17} \\
\midrule

\multirow{2}{*}{\textbf{Kimi} (3B/48B)}
& LoRA        & 0.27 & 94.33 & 79.24 & \textbf{91.04} & 97.35 & 75.14 & 87.40 & 90.59 & 54.15 & 83.66 \\
& \cellcolor{ourscolor}\textbf{\system{}}
& \cellcolor{ourscolor} 0.27
& \cellcolor{ourscolor}\textbf{94.82} & \cellcolor{ourscolor}\textbf{81.14}
& \cellcolor{ourscolor}90.36 & \cellcolor{ourscolor}\textbf{97.43}
& \cellcolor{ourscolor}\textbf{76.30} & \cellcolor{ourscolor}\textbf{88.00}
& \cellcolor{ourscolor}\textbf{90.97} & \cellcolor{ourscolor}\textbf{54.45}
& \cellcolor{ourscolor}\textbf{84.18} \\

\bottomrule
\end{tabular}
}
\caption{
{
Commonsense reasoning comparison between LoRA and \system{} across four MoE backbones. Results for OLMoE, Qwen1.5-MoE, and Moonlight are from Table~\ref{tab:main_results}; Kimi-Linear-48B-A3B is added here. \system{} achieves higher average accuracy than LoRA on every backbone, including the largest one. The best result per backbone-row pair is in bold.}}
\label{tab:kimi_comparison}
\end{table*}

\section{Details for Louvain Method}
\label{app:louvain}

\begin{table*}[t]

\centering
\small
\resizebox{0.88\textwidth}{!}{
\begin{tabular}{lccccccccc}
\toprule
Group. & HS & WG & ARC-c & ARC-e & BoolQ & OBQA & PIQA & SIQA & Avg. \\
\midrule
Louvain & {92.24} & {72.85} & {70.88} & {86.21} & {68.04} & {77.33} & {83.93} & {53.41} & {75.61} \\
Thresh. & {92.17} & {71.77} & {71.05} & {86.66} & {67.94} & {77.07} & {84.21} & {53.15} & {75.50} \\
\bottomrule
\end{tabular}
}
\caption{
{
Comparison between Louvain-based grouping and fixed-threshold grouping on OLMoE commonsense reasoning. ``Thresh.'' denotes fixed-threshold grouping with $t=0.9$. Both variants use the same ACE pipeline and differ only in how CKA-based expert groups are formed.}}
\label{tab:threshold_grouping}
\end{table*}


\subsection{Grouping Implementation Details}
\label{app:grouping_details}
\paragraph{Probe set construction.}
We construct the probe set \(X_{\mathrm{probe}}\) exclusively from the
fine-tuning training split and do not use validation or test examples for
expert grouping. 
After the warm-up phase, We collect \(N=10{,}000\) non-padding token activations from randomly ordered packed training sequences.
The same probe set is used for all experts within a layer, so
the resulting CKA scores reflect adapter-induced representations under
identical inputs rather than differences in routing frequency. We keep the
probe set fixed across ACE variants and do not tune it on downstream validation
performance.

\paragraph{Layer-wise graph construction.}
We perform expert grouping independently for each MoE layer. Let
\(S^{(\ell)} \in \mathbb{R}^{E_\ell \times E_\ell}\) denote the CKA similarity
matrix among the \(E_\ell\) experts in layer \(\ell\). We first symmetrize
\(S^{(\ell)}\) and set its diagonal entries to zero to remove self-similarity.
We then construct a non-negative weighted adjacency matrix \(W^{(\ell)}\) as
\[
W^{(\ell)}_{ij} = \max\!\bigl(0,\, S^{(\ell)}_{ij} - b^{(\ell)}\bigr),
\]
where \(b^{(\ell)}\) is the median of the off-diagonal entries of
\(S^{(\ell)}\). This transformation suppresses weak pairwise similarities and
retains only expert relations that exceed the typical similarity level within
layer \(\ell\). The resulting graph is therefore a layer-specific weighted
graph whose nodes correspond to experts and whose edge weights encode CKA
similarity above the layer-specific baseline.
\subsection{Layer-wise Grouping with the Louvain Method}
\label{app:louvain_grouping}

Given the preprocessed graph \(W^{(\ell)}\), we apply the Louvain method
independently to each MoE layer. We optimize the generalized weighted
modularity
\[
Q = \frac{1}{2m}\sum_{i,j}
\left(
W_{ij} - \gamma \frac{k_i k_j}{2m}
\right)\mathbf{1}[z_i = z_j],
\]
where \(k_i = \sum_j W_{ij}\) is the weighted degree of expert \(i\),
\(2m = \sum_{i,j} W_{ij}\), \(z_i\) denotes the group assignment of expert
\(i\), and \(\gamma\) is the resolution parameter. In all experiments, we set
\(\gamma = 0.8\).

Starting from singleton groups, the Louvain method repeatedly considers moving
an expert \(i\) to one of its neighboring groups. After removing \(i\) from
its current group, the modularity gain of assigning it to a candidate group
$\mathcal{C}$ is proportional to
\[
\Delta Q(i \rightarrow \mathcal{C}) \propto
k_{i,\mathrm{in}}(\mathcal{C}) - \gamma \frac{k_i\,\mathrm{tot}(\mathcal{C})}{2m},
\]
where \(k_{i,\mathrm{in}}(\mathcal{C})\) denotes the total edge weight from expert \(i\)
to group $\mathcal{C}$, and \(\mathrm{tot}(\mathcal{C})\) is the sum of weighted degrees of the
experts currently assigned to $\mathcal{C}$. We therefore greedily move each expert to
the neighboring group that yields the largest positive increase in modularity;
if no positive gain is available, the expert remains in its current group.
Once no further positive local move is possible, the resulting groups are
collapsed into super-nodes and the same procedure is repeated on the coarsened
graph. This yields a set of expert groups
\(\{\mathcal{G}^{(\ell)}_g\}_{g=1}^{G_\ell}\) for each layer \(\ell\).

{

\subsection{Comparison with Fixed-Threshold Grouping}
\label{app:threshold_grouping}

The default ACE implementation forms expert groups by applying Louvain community detection to the CKA similarity graph. To examine whether ACE depends critically on Louvain grouping, we compare it with a fixed-threshold variant that uses the same CKA similarity matrix but replaces only the grouping rule.

For each layer \(\ell\) and projection type \(p\), let \(S^{(\ell,p)} \in \mathbb{R}^{E \times E}\) denote the CKA similarity matrix across \(E\) experts. Fixed-threshold grouping constructs an expert graph using a threshold \(t\):

\[
\mathcal{E}^{(\ell,p)}_t =
\{(i,j) \mid i \neq j,\ S^{(\ell,p)}_{ij} \geq t \}.
\]

The expert groups are then defined as the connected components of \(\mathcal{G}^{(\ell,p)}_t=(\mathcal{V},\mathcal{E}^{(\ell,p)}_t)\), where $\mathcal{V}$ is the set of experts in layer $\ell$. We set \(t=0.9\). We use the same training setup as in Table~\ref{tab:main_results} and keep all other components of ACE unchanged, including rank consolidation, shared adapter updates, grouped adapter execution, and optimization hyperparameters.

Table~\ref{tab:threshold_grouping} shows that fixed-threshold grouping achieves performance close to Louvain grouping on OLMoE commonsense reasoning, with an average score of 75.50 compared to 75.61 for Louvain. This indicates that ACE does not depend critically on a specific grouping algorithm, as long as functionally similar experts are consolidated. However, Louvain slightly outperforms the fixed-threshold variant and avoids manually selecting a global threshold. We therefore use Louvain grouping as the default strategy.
}

\section{Implementation and Profiling of Grouped Adapter Execution}

\subsection{Implementation Details}
\label{sub:acc_scope}

Section~\ref{subsec:accel_adapter} presents grouped execution for a representative adapted linear transformation and omits projection indices for clarity. In practice, however, LoRA can be attached independently to the up, gate, and down projections of each expert, and the grouped execution argument applies only when the shared-input property holds.

For expert \(e\), the feed-forward network (FFN) can be written as
\[
u_e = (W_e^{\mathrm{up}} + \Delta W_e^{\mathrm{up}})x,
\]
\[
v_e = (W_e^{\mathrm{gate}} + \Delta W_e^{\mathrm{gate}})x,
\]
\[
h_e = \phi(v_e)\odot u_e,
\]
\[
y_e = (W_e^{\mathrm{down}} + \Delta W_e^{\mathrm{down}})h_e,
\]
where \(x\) is the routed token representation and \(\phi(\cdot)\) denotes the gating nonlinearity. This gated-FFN (SwiGLU-style) form is the architecture used by all MoE backbones in our experiments. The key distinction is that the up and gate projections both receive the same pre-expert input \(x\), whereas the down projection receives \(h_e\), which is already expert-specific because it depends on the outputs of the expert's own up/gate branches.

This distinction determines where grouped execution is effective. After consolidation, the shared adapter corrections for the up and gate projections can be computed once per active group,
\[
c_g^{\mathrm{up}}(x)=\Delta W_g^{\mathrm{up,share}}x,
\]
\[
c_g^{\mathrm{gate}}(x)=\Delta W_g^{\mathrm{gate,share}}x,
\]
using grouped GEMM, and then scattered to all routed experts assigned to that group. The same reuse is not available for the down projection. Even when the routed token is the same, the intermediate state \(h_e\) differs across experts, so evaluating a down-side adapter still requires expert-specific inputs. Recovering the same acceleration for the down projection would require additional adapter merging or weight materialization, which introduces nontrivial overhead. For this reason, our accelerated grouped adapter execution is implemented only for the up and gate projections.

{
\ifbluemode\color{accentblue}\fi

\subsection{Fine-Grained Runtime Profiling}
\label{app:runtime_profile}

To identify the source of the runtime improvement from grouped adapter
execution, we profile expert-wise LoRA and \system{} on OLMoE under
matched input batches, routing, trainable-parameter budgets, and optimizer
settings. Results are mean $\pm$ standard deviation over five runs, each
with 20 warm-up and 100 measured steps.

\begin{table}[t]
\centering
\footnotesize
\renewcommand{\arraystretch}{1.08}
\setlength{\tabcolsep}{3.5pt}

\resizebox{\columnwidth}{!}{%
\begin{tabular}{lccc}
\toprule
\textbf{Component}
& \textbf{LoRA (ms)}
& \textbf{\system{} (ms)}
& \textbf{Reduction} \\
\midrule
Total step
& $3010.38$ \std{33.86}
& $1688.13$ \std{9.31}
& 43.92\% \\
Forward
& $1351.95$ \std{17.20}
& $845.43$ \std{6.49}
& 37.47\% \\
Adapter forward
& $517.48$ \std{3.14}
& $56.09$ \std{0.28}
& \textbf{89.16\%} \\
Backward
& $1619.53$ \std{21.18}
& $809.91$ \std{3.30}
& 49.99\% \\
Optimizer
& $24.35$ \std{0.66}
& $21.88$ \std{0.25}
& -- \\
Router linear
& $0.330$ \std{0.001}
& $0.330$ \std{0.001}
& -- \\
\bottomrule
\end{tabular}%
}

\caption{Fine-grained runtime profiling of expert-wise LoRA and
\system{} with grouped adapter execution on OLMoE. Adapter forward is
a subrange of the forward pass.}
\label{tab:runtime_profile}
\end{table}

Grouped execution reduces post-consolidation step time by 43.92\%,
with the largest reduction occurring in adapter forward execution
(89.16\%). The adapter-forward reduction accounts for approximately
91\% of the measured forward-time reduction, confirming that the
efficiency gain is concentrated in the fragmented adapter computation
targeted by \system{}.

These synchronized profiling measurements are intended for
component-level attribution and are therefore not directly comparable
to the end-to-end wall-clock measurements in Table~\ref{tab:training_efficiency},
which additionally include data-pipeline and framework overhead.
}

\begin{figure}[t]
\centering
\includegraphics[width=0.49\columnwidth]{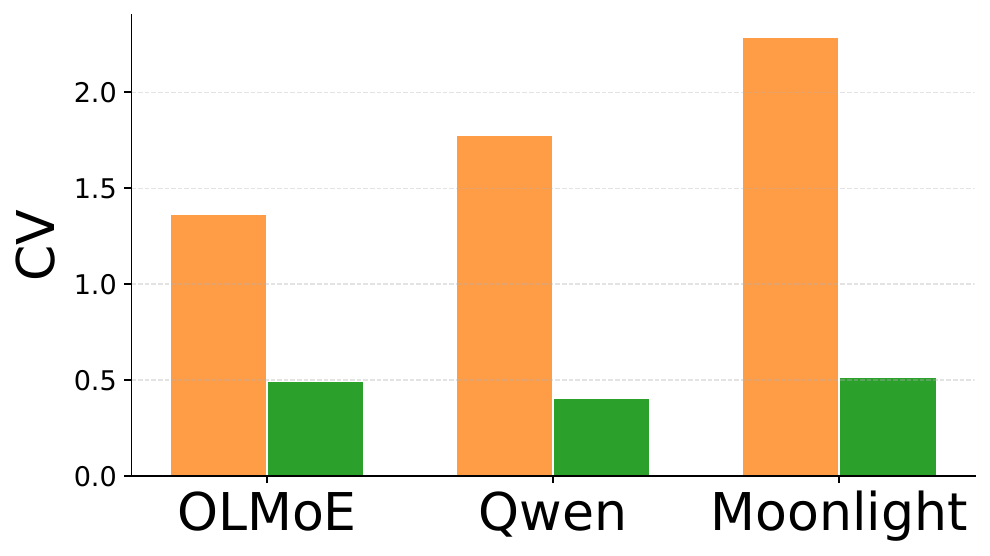}
\includegraphics[width=0.49\columnwidth]{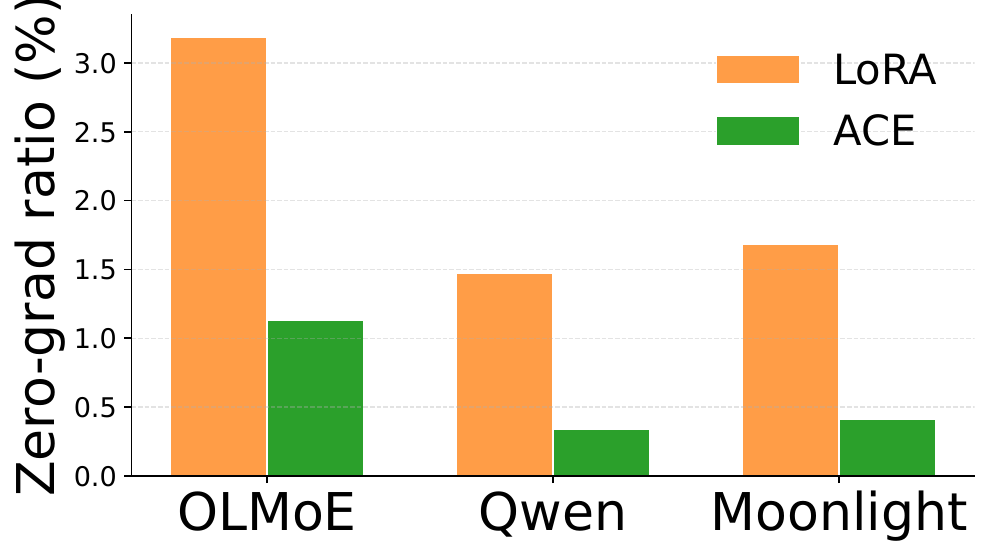}
\caption{Gradient-supervision fragmentation for gate-projection LoRA adapters across three MoE backbones. 
\system{} reduces both gradient-norm imbalance and the zero-gradient ratio relative to expert-wise LoRA.}
\label{fig:gate_grad_frag}
\end{figure}

{

\section{Gradient Supervision Analysis Details}
\label{app:grad_gate}

We provide additional details for the gradient-supervision fragmentation analysis in Section~\ref{sec:fragmentation_mitigation}. 
We conduct this analysis under the commonsense fine-tuning setting, using Commonsense-170K as the training data.
For each backbone and method, we measure LoRA-adapter gradients over 50 matched training steps, using the post-consolidation phase for \system{}.
We compute the statistics separately for each MoE layer and then average across layers.

For each expert adapter, we first compute the gradient norm of its LoRA parameters. 
We then measure two quantities. 
The first is the coefficient of variation (CV) of per-expert gradient norms within the same MoE layer,
\[
\mathrm{CV}
=
\frac{\mathrm{Std}_{e}\left(\|\nabla \Delta W_e\|\right)}
{\mathrm{Mean}_{e}\left(\|\nabla \Delta W_e\|\right)} ,
\]
which captures how unevenly gradient signal is distributed across experts. 
The second is the zero-gradient ratio, defined as the fraction of LoRA adapter rows whose gradients are exactly zero,
\[
\mathrm{ZeroGrad}
=
\frac{\#\{\text{adapter rows with zero gradient}\}}
{\#\{\text{adapter rows}\}} .
\]
A lower CV indicates more balanced supervision across experts, and a lower zero-gradient ratio indicates that fewer adapter parameters are left unsupervised at each step.

Figure~\ref{fig:gate_grad_frag} reports the same analysis for gate-projection adapters. 
Consistent with the up-projection results in Figure~\ref{fig:grad_frag}, \system{} reduces both gradient imbalance and zero-gradient ratios across MoE backbones. 
This supports the conclusion that shared-group updates densify supervision not only for up projections but also for gate projections.
}

\section{Additional Training Efficiency Results}
\label{app:efficiency}

Table~\ref{tab:appendix_method_efficiency} reports method-level training time and peak memory across MoE backbones. 
All methods are evaluated under the same effective batch size within each backbone. 
When a method requires a smaller per-device batch size due to memory constraints, gradient accumulation is adjusted to keep the effective batch size fixed. 
We report wall-clock training time over 3 epochs and peak GPU memory. 
Means and standard deviations are computed over available seeds(0, 1, 2).

\begin{table}[t]

\centering
\small
\setlength{\tabcolsep}{6pt}
\renewcommand{\arraystretch}{1.12}
\resizebox{\columnwidth}{!}{
\begin{tabular}{l l c c}
\toprule
\textbf{Model} & \textbf{Method} & \textbf{Train Time (hr)} & \textbf{Peak Memory (GB)} \\
\midrule

\multirow{5}{*}{Moonlight}
& LoRA    & 23.75 $\pm$ 0.60 & 48.80 $\pm$ 0.00 \\
& AdaLoRA & 43.84 $\pm$ 2.05 & 50.07 $\pm$ 0.62 \\
& DoRA    & 41.10 $\pm$ 0.95 & 80.88 $\pm$ 0.02 \\
& PiSSA   & 23.64 $\pm$ 0.33 & 48.80 $\pm$ 0.00 \\
& \textbf{\system{} (ours)} & \textbf{16.77 $\pm$ 0.23} & 48.80 $\pm$ 0.00 \\

\midrule

\multirow{5}{*}{Qwen}
& LoRA    & 6.18 $\pm$ 0.05 & 89.90 $\pm$ 0.01 \\
& AdaLoRA & 10.90 $\pm$ 0.34 & 90.58 $\pm$ 0.00 \\
& DoRA    & 18.56 $\pm$ 0.06 & 96.57 $\pm$ 0.01 \\
& PiSSA   & 6.25 $\pm$ 0.09 & 89.84 $\pm$ 0.01 \\
& \textbf{\system{} (ours)} & \textbf{4.19 $\pm$ 0.02} & 89.84 $\pm$ 0.01 \\

\midrule

\multirow{5}{*}{OLMoE}
& LoRA    & 2.47 $\pm$ 0.03 & 98.21 $\pm$ 0.01 \\
& AdaLoRA & 3.90 $\pm$ 0.05 & 98.54 $\pm$ 0.00 \\
& DoRA    & 7.36 $\pm$ 0.45 & 85.10 $\pm$ 0.00 \\
& PiSSA   & 2.40 $\pm$ 0.03 & 98.22 $\pm$ 0.00 \\
& \textbf{\system{} (ours)} & \textbf{1.89 $\pm$ 0.02} & 98.18 $\pm$ 0.02 \\

\midrule

\multirow{2}{*}{Kimi-Linear}
& LoRA    & 87.39 $\pm$ 0.80 & 118.22 $\pm$ 0.00 \\
& \textbf{\system{} (ours)} & \textbf{60.41 $\pm$ 0.38} & 118.22 $\pm$ 0.00 \\

\bottomrule
\end{tabular}
}
\caption{
Additional training efficiency results across PEFT methods and MoE backbones. 
Training time is reported as wall-clock hours over 3 epochs, and peak memory is reported in GB. 
Means and standard deviations are computed over three seeds (0, 1, and 2). 
All methods use the same effective batch size within each backbone; methods with higher memory requirements may use smaller per-device batches with additional gradient accumulation. 
For Kimi-Linear, we report the available LoRA comparison.}

\label{tab:appendix_method_efficiency}
\end{table}

\begin{table*}[!ht]

\centering
\small
\setlength{\tabcolsep}{4pt}
\renewcommand{\arraystretch}{1.15}
\resizebox{\textwidth}{!}{
\begin{tabular}{l c c c c c c c c c}
\toprule
\textbf{Model} & \textbf{$E$} & \textbf{Layers} & \textbf{Total Experts}
& \textbf{Warm-up Overhead (s)} & \textbf{CKA (s)} & \textbf{Louvain (s)} & \textbf{Consolidation (s)}
& \textbf{Total Overhead (\% of ACE)}
& \textbf{ACE Speedup over LoRA} \\
\midrule
OLMoE-1B-7B          & 64  & 16 & 1{,}024 & 0.00 & 123.90 & 0.12 & 0.37 & 0.035\,hr (1.83\%) & 1.31$\times$ \\
Qwen1.5-MoE-A2.7B    & 60  & 24 & 1{,}440 & 0.00 & 275.17 & 0.15 & 0.94 & 0.077\,hr (1.83\%) & 1.48$\times$ \\
Moonlight-16B-A3B    & 64  & 27 & 1{,}728 & 0.00 & 328.89 & 0.19 & 0.56 & 0.092\,hr (0.55\%) & 1.42$\times$ \\
Kimi-Linear-48B-A3B  & 256 & 27 & 6{,}912 & 0.00 & 2393.11 & 0.79 & 3.19 & 0.666\,hr (1.10\%) & 1.45$\times$ \\
\bottomrule
\end{tabular}
}
\caption{
Per-stage breakdown of the \system{} consolidation pipeline. The three stages (pairwise CKA computation, Louvain community detection, and adapter consolidation) are reported in seconds. Total overhead is the sum of the three stages relative to total ACE training time. The warm-up phase itself adds no extra cost, as it is identical to standard LoRA training. All models are trained for 3 epochs, following the configuration of Table~\ref{tab:main_results}. Shared experts, which are not included in the ACE grouping procedure,
are excluded from the expert counts.}
\label{tab:overhead_breakdown}
\end{table*}
{

\section{Overhead Analysis of the Consolidation Pipeline}
\label{app:overhead_breakdown}
The consolidation pipeline of \system{} introduces a one-time preprocessing cost that consists of three stages: (i) CKA computation across all expert pairs within each MoE layer, (ii) Louvain community detection on the resulting similarity graph, and (iii) adapter consolidation (CKA-center anchor selection, shared high-rank adapter initialization, and optimizer-state reallocation). Note that the warm-up phase itself contributes \emph{no} additional cost: during $T_{\text{init}}$ steps, \system{} simply trains expert-wise LoRA adapters identically to the LoRA baseline, so the warm-up time is already accounted for in the total training time and is not counted as overhead.
The dominant scaling term is the pairwise CKA computation, which has $\mathcal{O}(E^2)$ complexity per layer, where $E$ is the number of experts.
We report a per-stage breakdown across all four MoE backbones to characterize how this overhead grows with $E$ and how much it costs relative to total training time.

Table~\ref{tab:overhead_breakdown} shows three observations.
First, the consolidation pipeline accounts for at most 1.83\% of total ACE training time across all evaluated backbones, despite the $\mathcal{O}(E^2)$ scaling of pairwise CKA.
Second, CKA computation dominates the overhead, consistently accounting for the overwhelming majority of the one-time cost across all backbones, while Louvain and adapter consolidation remain almost flat and contribute a negligible fraction even at the largest scale. The cost of CKA computation is controllable via the size of the probe set used to compute the similarity matrices. A smaller probe set proportionally reduces CKA runtime, reducing overhead on larger MoE backbones.
Third, the overhead is more than offset by per-step efficiency gains from grouped adapter execution: \system{} achieves at least a $1.31\times$ end-to-end speedup over LoRA on every backbone.
}

\section{Optimization Dynamics}

\begin{figure}[htbp]
\centering
\includegraphics[width=\columnwidth]{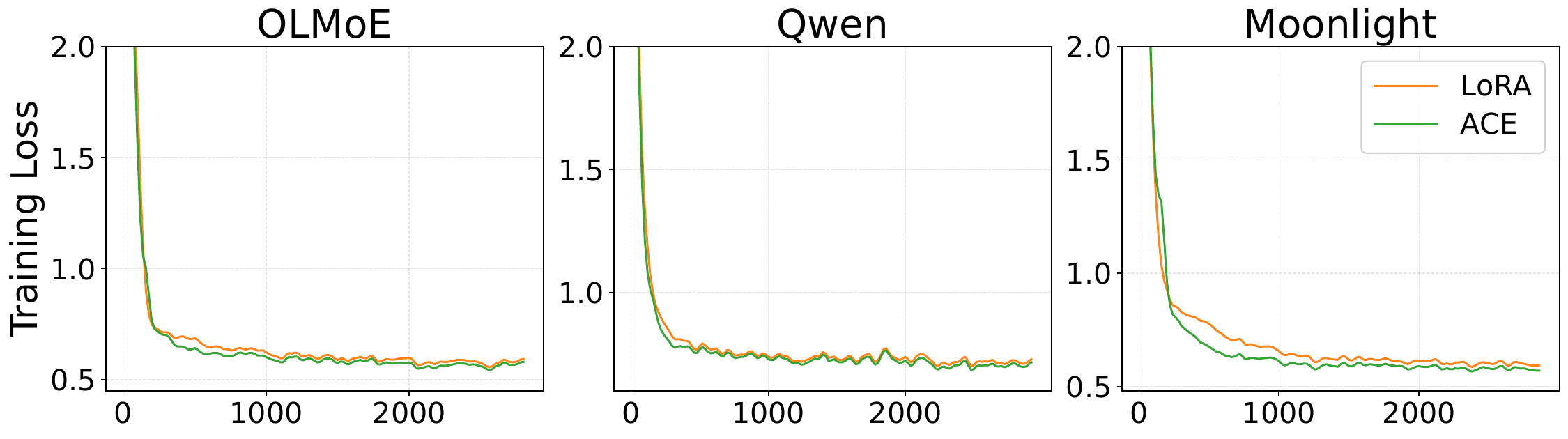}
\vspace{-2mm}
\caption{
{Training loss curves on OLMoE, Qwen, Moonlight with Commonsense.}
}
\label{fig:loss_olmoe}
\vspace{-3mm}
\end{figure}

\begin{figure}[htbp]
\centering
\includegraphics[width=\columnwidth]{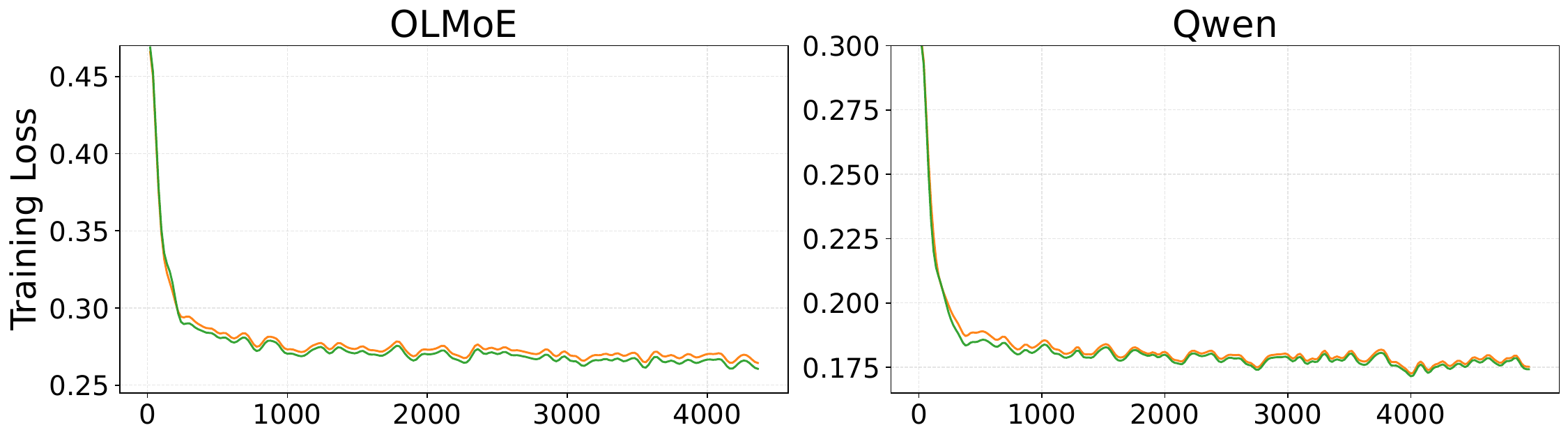}
\vspace{-2mm}
\caption{
{Training loss curves on OLMoE, Qwen with MetaMathQA.}
}
\label{fig:loss_qwen}
\vspace{-3mm}
\end{figure}

\begin{figure}[h!]
\centering
\includegraphics[width=\columnwidth]{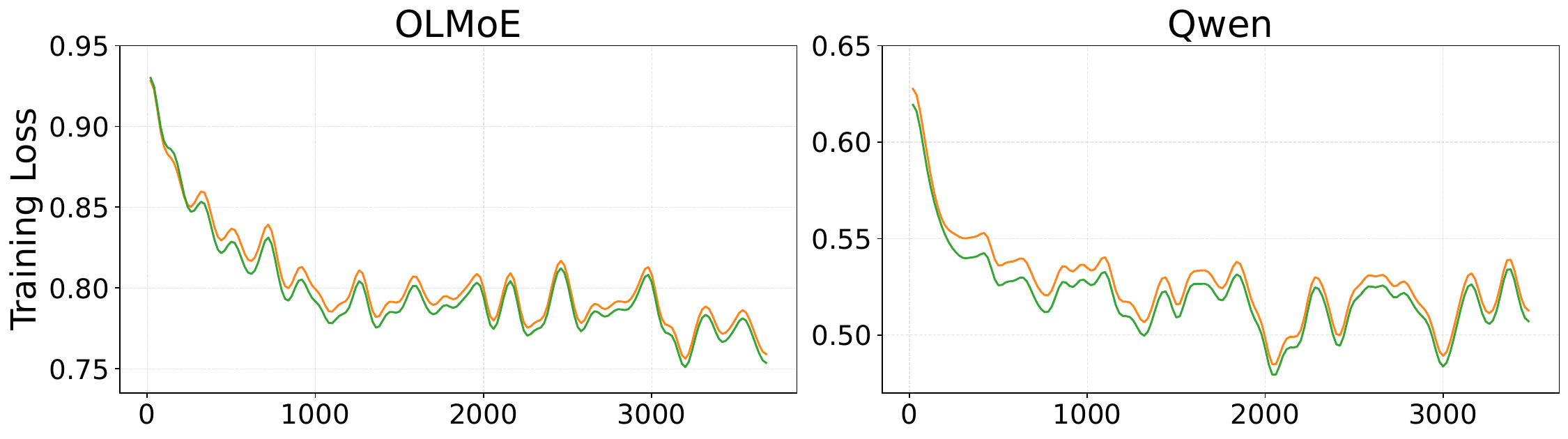}
\vspace{-2mm}
\caption{
{Training loss curves on OLMoE, Qwen with CodeFeedback.}
}
\label{fig:loss_moonlight}
\vspace{-3mm}
\end{figure}

\begin{figure}[h!]
\centering
\includegraphics[width=\columnwidth]{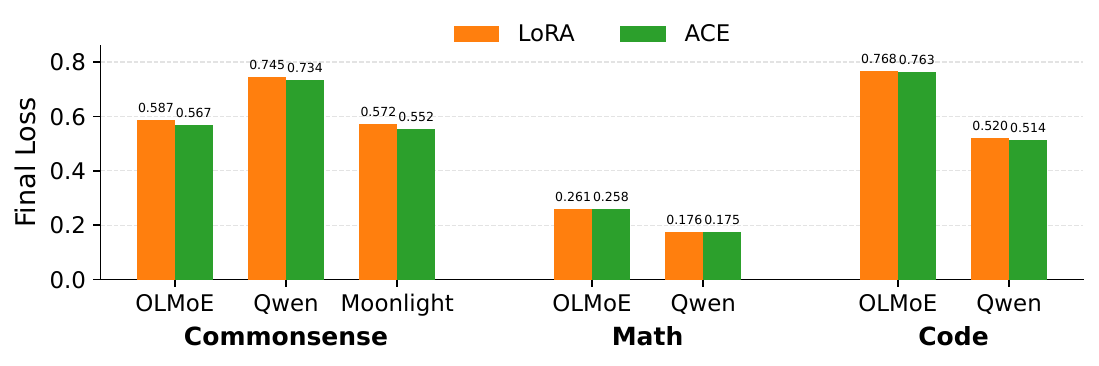}
\vspace{-2mm}
\caption{
{
Final training loss of LoRA and \system{} across datasets and MoE backbones. Lower values indicate better convergence.
}
}
\label{fig:final_loss_by_dataset}
\vspace{-3mm}
\end{figure}

{As shown in Figure \ref{fig:loss_olmoe}, \ref{fig:loss_qwen}, and \ref{fig:loss_moonlight}, after the merge step, \system{} demonstrates a steeper decrease in training loss compared to LoRA, suggesting that consolidating functionally similar adapters leads to more effective optimization dynamics. This improvement disappears immediately after merging due to initialization overhead.  After stabilization, the improvement appears and persists throughout training. Consistently, Figure~\ref{fig:final_loss_by_dataset} shows that \system{} reaches lower final training loss than LoRA across all evaluated dataset--backbone pairs, indicating that the post-merge optimization benefit carries through to convergence. For analysis, we report training loss curves after applying one-dimensional Gaussian smoothing to the plotted loss values. The Gaussian smoothing scale was set separately for each dataset: $\sigma$ = 1.5 for MetaMathQA, $\sigma$ = 2.0 for CodeFeedback, and $\sigma$ = 1.0 for Commonsense Reasoning.}

\section{LLM Usage}
We used an LLM to refine the writing, improve grammatical accuracy,
and assist with code organization and cleanup.